\documentclass[preprint,3p,times,twocolumn]{elsarticle}

\usepackage{amsmath}
\usepackage{amssymb}
\usepackage[nolist]{acronym}
\usepackage{bm}
\usepackage{booktabs}
\usepackage{color}
\usepackage[]{graphicx}
\usepackage{url}
\usepackage[caption=false, font=footnotesize]{subfig}
\usepackage{dblfloatfix}
\usepackage{multirow}
\usepackage{balance} %

\usepackage{relsize}
\usepackage{standalone} %
\usepackage{tikz}
\usetikzlibrary{shapes, arrows, arrows.meta,  positioning, shapes.geometric}

\usepackage{scrextend}

\graphicspath{{./figures/}}

\newcommand{\revise}[1]{{\color{black}#1}}
\newacro{dof}[DoF]{Degrees of Freedom}
\newacro{dl}[DL]{Deep Learning}
\newacro{ml}[ML]{Machine Learning}
\newacro{ds}[DS]{Dynamical System}
\newacro{dvs}[DVS]{Direct \ac{vs}}
\newacro{gmr}[GMR]{Gaussian Mixture Regression}
\newacro{ik}[IK]{Inverse Kinematics}
\newacro{il}[IL]{Imitation Learning}
\newacro{iil}[IIL]{Image-based \ac{il}}
\newacro{ilvs}[ILVS]{Imitation Learning Visual Servoing}
\newacro{ildvs}[IL-DVS]{Imitation Learning-based Direct Visual Servoing}
\newacro{nn}[NN]{artificial Neural Network}
\newacro{vs}[VS]{Visual Servoing}
\newacro{rds}[RDS]{Reshaped \acl{ds}}
\newacro{clfdm}[CLF-DM]{Control Lyapunov Function-based Dynamic Movements}
\newacro{clf}[CLF]{Control Lyapunov Function}
\newacro{fdm}[FDM]{Fast Diffeomorphic Matching}
\newacro{dmp}[DMP]{Dynamic Movement Primitive}
\newacro{seds}[SEDS]{Stable Estimator of Dynamical Systems}

\newcommand{\codeurl}{\url{https://github.com/sayantanauddy/il-dvs}}

\journal{Robotics and Autonomous Systems}

\usepackage{etoolbox}
\makeatletter
\patchcmd{\ps@pprintTitle}
  {Preprint submitted to}
  {To appear in}
  {}{}
\makeatother

\begin{document}
\begin{frontmatter}

\title{\LARGE \textbf{Imitation Learning-based Direct Visual Servoing\\using the Large Projection Formulation
}}
\author[1]{Sayantan Auddy\corref{cor1}\corref{shared}}
\ead{sayantan.auddy@student.uibk.ac.at}
\author[2]{Antonio Paolillo\corref{shared}}
\ead{antonio.paolillo@idsia.ch}
\author[1,3]{Justus Piater}
\ead{justus.piater@uibk.ac.at}
\author[4]{Matteo Saveriano}
\ead{matteo.saveriano@unitn.it}
\cortext[cor1]{Corresponding author.}
\cortext[shared]{These authors contributed equally.}
\address[1]{Department of Computer Science, University of Innsbruck, Innsbruck, Austria}
\address[2]{Dalle Molle Institute for Artificial Intelligence (IDSIA), USI-SUPSI, Lugano, Switzerland}
\address[3]{Digital Science Center (DiSC), University of Innsbruck, Innsbruck, Austria}
\address[4]{Department of Industrial Engineering, University of Trento, Trento, Italy}

\begin{abstract}
Today robots must be safe, versatile, and user-friendly to operate in %
unstructured and human-populated environments. Dynamical system-based imitation learning enables robots to perform complex tasks stably and without explicit programming, greatly simplifying their real-world deployment. To exploit the full potential of these systems it is crucial to implement closed loops that use visual feedback. Vision permits to cope with environmental changes, but is complex to handle due to the high dimension of the image space. This study introduces a dynamical system-based imitation learning for direct visual servoing. It leverages off-the-shelf deep learning-based perception \revise{modules} to extract robust features from the raw input image, and an imitation learning strategy to execute sophisticated robot motions. The learning blocks are integrated using the large projection task priority formulation. As demonstrated through extensive experimental analysis, the proposed method realizes complex tasks with a robotic manipulator.
\end{abstract}


\end{frontmatter}

\begin{figure}[ht!]
\centering
\frame{\includegraphics[trim={10cm 1.5cm 10cm 0},clip,width=0.7\columnwidth]{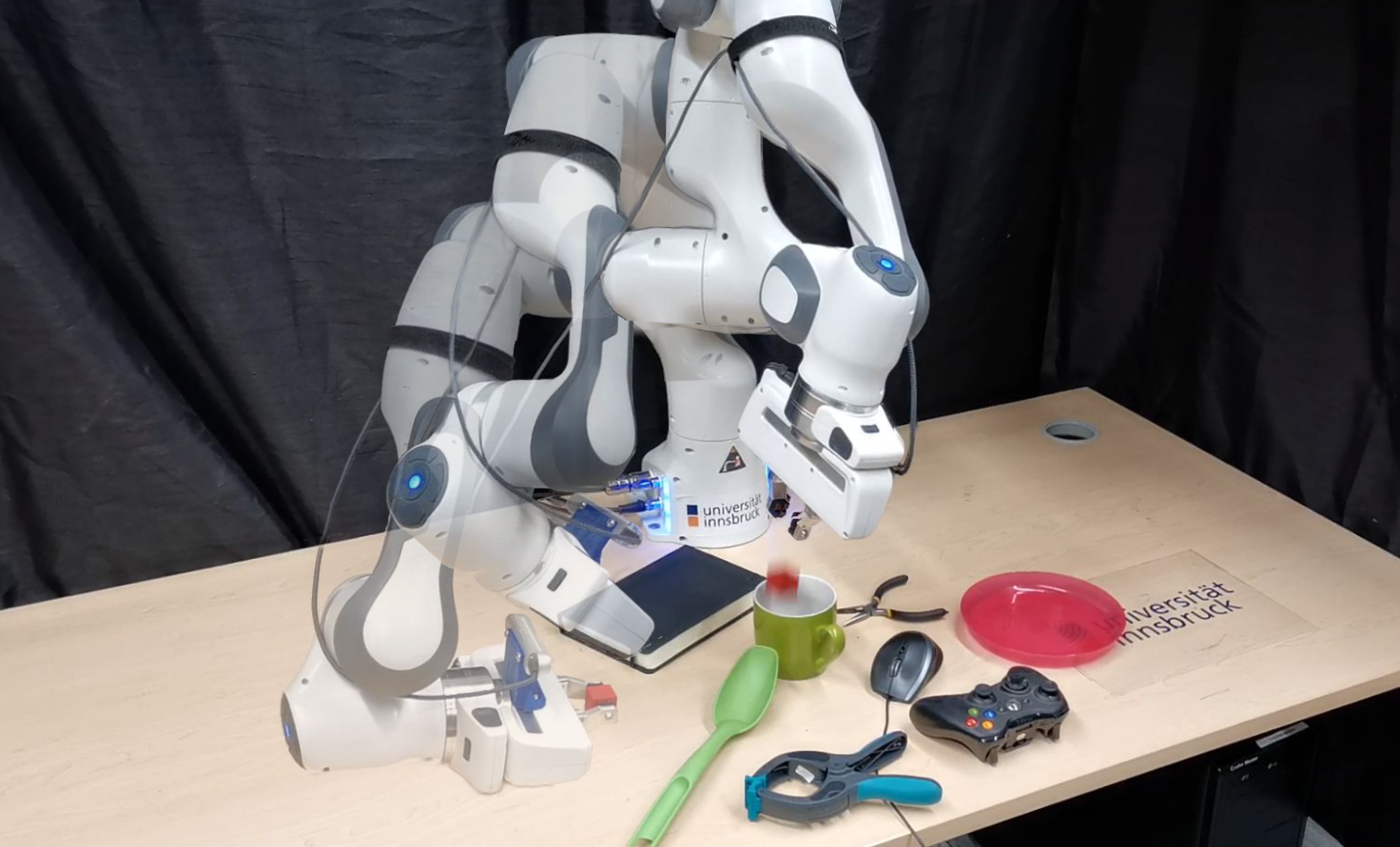}}
\caption{Our work combines off-the-shelf deep learning strategies to detect objects in the clutter, and imitation learning to realize complicated trajectories, e.g., dropping a cube into a cup on an untidy table. The large projection formulation combines the two machine learning components and ensures convergence to a given target.}
\label{fig:intro}
\end{figure}
\section{Introduction}\label{sec:intro}
Modern robots must be accessible to everyone, as they are rapidly spreading in everyday life environments, from industries~\cite{grau2020robots} to hotels~\cite{Choi:jhmm:2020} and hospitals~\cite{Gonzalez:as:2021}.
It is expected that a growing number of inexperienced end-users, like children, patients, or elderly people, ask for robots with easy-to-use, friendly, and modular interfaces, endowed with adaptive skills.
To meet these requirements, recent advancements in robotics demonstrate the great potential of Machine Learning (\ac{ml}).

From a control perspective, ease of use and adaptability can be obtained with vision-based \ac{il}.
On one side, \ac{il} allows one to easily implement robotic tasks without specific codes~\cite{Argall:ras:2009,Billard:handbook:2008}, but by simply following a few demonstrations.
\acp{ds} handle the imitation strategy by keeping the stability properties of classical controllers.
Such approaches successfully generate complex kinematic motion from previous demonstrations, see~\cite{Khansari:tro:2011, Khansari:ras:2014,Perrin:scl:2016, Urain:iros:2020}.
On the other, vision-based control like \ac{vs}~\cite{Chaumette:ram:2006,Chaumette:ram:2007} generates robot behaviors from exteroceptive information, thus taking into account possible changes in the environment.
In recent work~\cite{Paolillo:icra:2022,Paolillo:icra:2023}, \ac{ds}-based \ac{il} and \ac{vs} are combined to realize the so-called \ac{ilvs}. 
Such a scheme provides dual benefits: (i) additional and complex skills can be imitated (and not explicitly coded) in the \ac{vs} law; (ii) the use of vision enables adaptive imitation strategies.
In this way, the limitations of the original law are overcome by leveraging the information of the demonstrations, e.g., for realizing a visual tracker without an explicit target motion estimator~\cite{Felici:hfr:2023}.

From a perception point of view, it would be desirable to have modular and transferable blocks that could be easily adapted to the specificity of the deployed robot and the considered task.
In this context, \ac{dl} has been demonstrated to outperform classical computer vision methods~\cite{OMahony:cvc:2020}, also in the robotics domain~\cite{Goodfellow:book:2016}.
Indeed, many approaches based on \ac{dl} show great performance in detecting and tracking complex objects, e.g., the well-known You Only Look Once (YOLO) algorithm~\cite{Redmon:cvpr:2016, jiang2022review}.
These approaches exhibit robustness, versatility, and even generalization capability, and can solve complex perception tasks in the robotic domain~\cite{Goodfellow:book:2016,OMahony:cvc:2020}.
However, sporadic misinterpretations of the raw sensor data, or hallucinations, typical of \ac{dl} approaches~\cite{sahoo2024unveiling}, could be disruptive in a closed-loop control scheme. 
Indeed, the measurement accuracy required by a precise control action sets severe requirements, which might be difficult to fulfill by \ac{dl}-based perception.
Specific systems like YOLO, for example, are very reliable in recognizing an object and its position in the camera field of view. 
However, they fall short in estimating its right orientation, preventing full 6D pose regulation or tracking.

This work aims to realize a robust, modular, stable yet simple \ac{vs} scheme, taking the good from both data-driven and model-based approaches.
We propose a \ac{vs} architecture that leverages \ac{il} and \ac{dl} paradigms to exploit the potential of state-of-the-art detectors and overcome their limitations in control loops.
More in detail, we propose to use an off-the-shelf \ac{dl}-based detector to obtain a rough but robust visual feedback and refine it by performing \ac{il} from a few demonstrations of the desired \ac{vs} behavior.
The learning components are combined in a formal model-based control structure. %
In this way, we target robotic applications (like dropping a sugar cube in a cup of tea on an untidy table, as exemplified in Fig.~\ref{fig:intro}) proposing an easy solution to complex perception problems, and simple generation of complicated trajectories.

The remainder of the paper is organized as follows. 
Sec.~\ref{sec:related_work} discusses the related literature, whereas the technical background of our work is presented in Sec.~\ref{sec:background}. 
Our method and the experimental setup used to validate it are detailed in Sec.~\ref{sec:approach} and Sec.~\ref{sec:ext_setup}, respectively.
The approach is validated with an extensive experimental analysis, whose results are presented in Sec.~\ref{sec:results}.
Section~\ref{sec:conclusion} concludes the paper with final remarks.

\section{Related work}\label{sec:related_work}

In our vision of adaptive and easy-to-use robots, we must design visual controllers that are straightforward to deploy. 
In practice, we aim to avoid specific coding to (i) extract the required feedback from dense images and (ii) generate sophisticated trajectories.

Impressive off-the-shelf software releases, e.g., YOLO~\cite{Redmon:cvpr:2016}, have been shown to detect objects robustly.
It is worth mentioning the large body of work that aims at estimating from vision the target object's pose, see, e.g.,~\cite{Li:eccv:2018,Deng:icra:2020,Nava:ral:2021,Nava:ral:2022}, which can also serve as a control feedback. 
However, all these approaches implement \emph{standalone} perception systems, i.e., they are unaware of the underlying control structure, and their output might not be accurate enough for control purposes.
Specific pose estimators for vision-based control have also been proposed~\cite{Saxena:icra:2017,Yu:iros:2019,Bateux:icra:2018,durdevic2020deep, vitiello2023one}, but these approaches are sensitive to the operating conditions and usually need intense retraining to operate in different scenarios. 
Furthermore, pure perception algorithms delegate the generation of sophisticated trajectories to the control block.

One possibility consists of coupling perception and control together in end-to-end learning fashion~\cite{Felton:icra:2021,Puang:iros:2020,Levine:jmlr:2016}.
\revise{However, end-to-end methods cannot guarantee stability properties and robustness to disturbances. This is particularly challenging if the robot needs to operate in dynamically changing environments and/or close to the human.}
Coarse-to-fine imitation learning~\cite{Johns:icra:2021} combines closed- and open-loop execution to perform complex manipulation tasks.  
One way to ensure stability is to maintain the formal structure of the visual controller, e.g., preserving the \ac{vs} formalism.
In the context of our work, \ac{dvs} is particularly interesting because it implements \ac{vs} using direct image measurement, avoiding explicit feature extraction.
Examples are \ac{vs} schemes that use photometric moments~\cite{Bakthavatchalam:tro:2018}, pixel luminance~\cite{Collowet:tro:2011},
histograms~\cite{Bateux:ral:2017}, or Gaussian mixtures~\cite{Crombez:tro:2019} as control feedback.
However, in these approaches, the potential of \ac{dl} is not fully exploited.
In~\cite{Paolillo:iros:2022}, an \ac{nn} is trained using the knowledge of \ac{vs} to produce geometrically interpretable features.  
In~\cite{Felton:ral:2022}, an autoencoder is learned to reduce the dimensionality of the image space, and an interaction matrix is directly computed from the network using auto differentiation and utilized in a \ac{vs} law. 
Liu et al.~\cite{liu2022mgbm} simplify the YOLO \revise{architecture} %
to speed up object detection, while Luo et al.~\cite{luo2023robot} propose a top-down feature detection network, and both use the predicted features in a \ac{vs} scheme. 
\revise{These approaches perform well with \ac{vs} tasks from images without performing classical feature extraction but do not execute complex movements.
Our work, instead, provides a unified solution that generates sophisticated robot motion in addition to enhancing the robustness of the perception module.}

This paper \revise{presents} an approach to solve the perception problem and the generation of complex trajectories simultaneously, in the context of vision-based controllers. 
The proposed framework overcomes the limitations of the literature by proposing an \revise{\ac{ildvs}} strategy that integrates off-the-shelf DL-based perception \revise{modules} with \ac{il} in a control theoretic framework.

\section{Background}\label{sec:background}

Image-based \ac{vs}~\cite{Chaumette:ram:2006} is a more than twenty-year-old established technique to regulate a camera to a desired pose through visual information.
The most basic law computes 6D velocity commands as $\bm{v}=-\lambda \hat{\bm{L}}^+ \bm{e}$~\cite{Chaumette:ram:2006}, zeroing an error $\bm{e}\in\mathbb{R}^f$ defined on the image; $\lambda$ is a scalar gain; $\hat{\bm{L}}^+\in\mathbb{R}^{f\times 6}$ is the pseudo-inverse of the so-called interaction matrix, relating the camera velocity to the time derivative of the visual feedback. \revise{The interaction matrix normally relies on the camera parameters that can be obtained through calibration procedures, and other information like the features' depth that needs to be estimated or approximated. Indeed,} the hat over the matrix indicates the approximation due to unknown $3$D parameters.

\begin{figure*}[t!]
\centering
\subfloat[\revise{Side view.}]{\includegraphics[width=0.6\columnwidth]{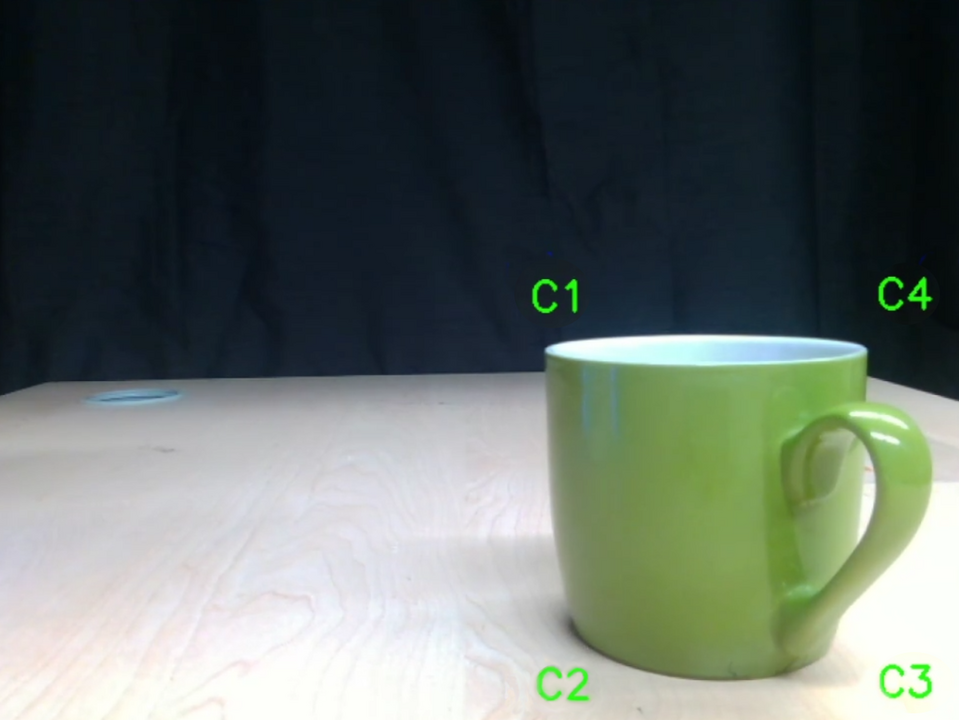}}
\quad
\subfloat[\revise{Oblique view.}]{\includegraphics[width=0.6\columnwidth]{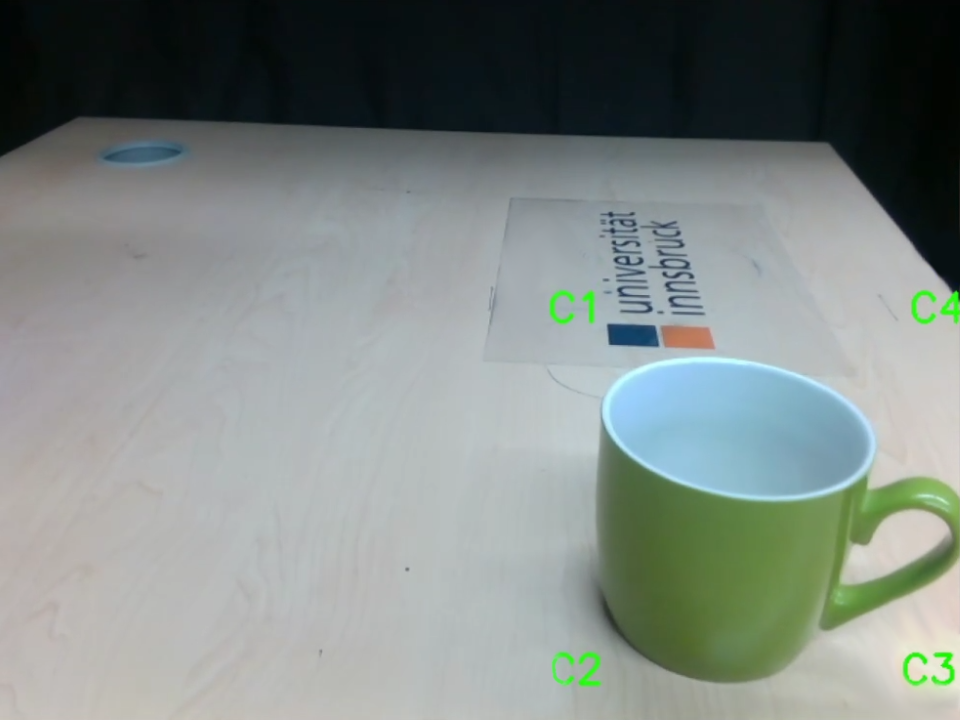}}
\quad
\subfloat[\revise{Top view.}]{\includegraphics[width=0.6\columnwidth]{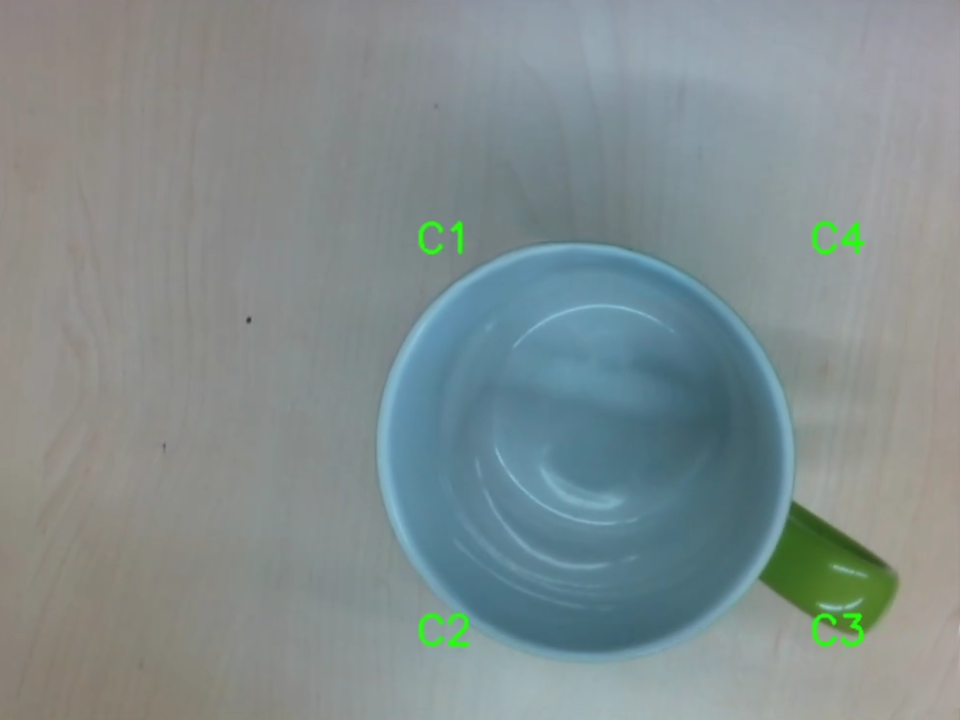}}
\caption[]{State-of-the-art \ac{dl}-based systems like YOLO can be used to detect the features of an object of interest on \revise{a} raw \revise{monocular} image robustly. Examples of features are the vertices (denoted with `C1', `C2', `C3', and `C4') of the bounding box detected around the image of \revise{a cup}. However, such detection systems fail to capture the correct object orientation. In the three snapshots, YOLO provides very similar feature values that correspond to \revise{three} very different relative camera-object orientations \revise{producing a side (a), oblique (b), and top view (c) of the cup}.
}
\label{fig:yolo_orientation}    
\end{figure*}

\ac{vs} can be executed together with other tasks, using the priority scheme established by the null-space projector~\cite{Siciliano:book:2009}:
\begin{equation}
    \bm{v}_e = -\lambda \hat{\bm{L}}^+ \bm{e} + \bm{P} \bm{\sigma}.
    \label{eq:vs}
\end{equation}
The matrix $\bm{P}=\bm{I}_6 - \hat{\bm{L}}^+ \hat{\bm{L}}$ is a null-space projector, and $\bm{\sigma}\in\mathbb{R}^6$ is the desired velocity realizing the secondary task.
The main limitation of~\eqref{eq:vs} is that, in normal working conditions, the dimension of the feedback is always greater or equal to the dimension of the task (i.e., $f\geq6$)~\cite{Chaumette:ram:2006}.
Under these circumstances, there is not much room in the null space of the primary task to execute any other secondary task.
Therefore, it has been proposed to use the norm of the error $\eta=\Vert \bm{e} \Vert$ \revise{in} the primary task~\cite{Marey:icra:2010} and to realize a prioritized control scheme in the following form 
\begin{equation}
    \bm{v}_\eta = -\lambda \eta \hat{\bm{L}}_\eta^+ + \bm{P}_\eta \, \bm{\sigma},
    \label{eq:vs_norm}
\end{equation}
\revise{where, similarly to the classic \ac{vs} scheme~\eqref{eq:vs}, $\bm{P}_\eta$ is a null-space projector and $\bm{\sigma}\in\mathbb{R}^6$ is the desired velocity realizing the secondary task.} The matrices of interest can be retrieved in closed form\revise{~\cite{Marey:icra:2010},  and are reported here for convenience:
\begin{equation*}
    \hat{\bm{L}}_\eta^+ = \eta \frac{\bm{\hat{L}}^\top \bm{e}}{\bm{e}^\top\bm{\hat{L}}\bm{\hat{L}}^\top\bm{e}}
    \quad\text{and}\quad
    \bm{P}_\eta = \bm{I}_6 - \frac{\bm{\hat{L}}^\top \bm{e}\,\bm{e}^\top\bm{\hat{L}}}{\bm{e}^\top\bm{\hat{L}}\bm{\hat{L}}^\top\bm{e}}.
\end{equation*}}%
Note that the control law~\eqref{eq:vs_norm}  drives the system towards the original desired behavior, as $\bm{e}\to \bm{0}$ for $\eta\to0$.
The operator $\bm{P}_\eta$ is called \emph{large projector} as the law~\eqref{eq:vs_norm} ensures enough room in the null space to achieve secondary tasks.
In practice, regulating the scalar norm instead of the vector error releases degrees of freedom during the transient for other secondary tasks.
Using simple Lyapunov arguments, it is possible to show that the law~\eqref{eq:vs_norm} is locally stable and that the secondary task does not impact the stability.
If the secondary velocities $\bm{\sigma}$ are not compatible with the primary task, they are simply ignored and not executed, as the effect of the construction of the projector operator~\cite{Siciliano:book:2009}.
\revise{In~\eqref{eq:vs_norm} the presence of singularities requires the use of a switching strategy} around $\eta = 0$:
\begin{equation}
    \bm{v} = \alpha(\eta)\,\bm{v}_\eta + \big(1 - \alpha(\eta)\big)\,\bm{v}_e
    \label{eq:vs_normal_and_norm}
\end{equation}
\revise{where $\alpha$ is a scalar variable smoothly changing from $1$ to $0$ in the vicinity of $\eta=0$, allowing a switch to the classical law~\eqref{eq:vs} from~\eqref{eq:vs_norm}~\cite{Marey:icra:2010}.}

This control system has been recently used to implement a stable \ac{il}~\cite{Paolillo:icra:2023} to realize complex \ac{vs} tasks.
In particular, the main task error $\eta$ stably drives the system towards steady-state convergence; the secondary task is used to imitate demonstrated velocities $\bm{\sigma}$ during the transient.
In this work, we use such a control structure to handle the output of a state-of-the-art \ac{dl}-based feature detector and overcome its limitations by leveraging the information of a few demonstrations, using the \ac{il} paradigm, as detailed in the following section.

\section{Approach}\label{sec:approach}

Our objective is to exploit the control redundancy offered by the large projector formalism~\eqref{eq:vs_norm} to integrate \ac{dl} and \ac{il} for direct \ac{vs}.
We exploit the great potential of state-of-the-art \ac{dl}-based object detectors, which are treated as an \revise{underlying raw} detector. %
Furthermore, our framework uses \ac{il} to look at previous demonstrations to overcome the limitations of \ac{dl}-based detectors.
The scheme further exploits \ac{il} to realize complex trajectories.
In this section, we explain in detail the \ac{dl} and \ac{il} learning components, and how they are combined using the large null-space projector control structure.
More in detail, we first present and formulate the perception problem in Sec.~\ref{sec:backbone}, which regards the limitation of the \revise{used} %
detectors in the control context; then, we present our solution to the problem introducing the imitation strategy (in Sec.~\ref{sec:imitation}) and the whole control scheme (Sec.~\ref{sec:control}) of our approach.

\subsection{DL-based detection %
and its limits}\label{sec:backbone}

The \ac{dl}-based %
detector is a pre-trained \ac{nn} that is fed with raw camera images and provides as output a measure of the object in the form of visual features
\begin{equation}
    \bm{f} = \bm{m}_{\bm{\theta}}(\bm{i}),
    \label{eq:model_backbone}
\end{equation}
where $\bm{i}\in\mathbb{R}^{3 w h}$ is a vectorized colored image with a size of $w \times h$ pixels, $\bm{m}$ is the \revise{detector model}, %
and $\bm{\theta}\in\mathbb{R}^p$ denotes the parameters of the pre-trained model; the output $\bm{f}\in\mathbb{R}^{m}$ contains features of the detected objects, such as the corners of its bounding box measured on the image (see Fig.~\ref{fig:yolo_orientation}).
Following the classic \ac{vs} rationale, such features are compared with a set of desired values, denoted with $\bm{f}^\ast$, to provide a measure of the visual error, from which the control action evolves.
Indeed, the desired set of features is obtained by the model fed with a reference desired image $\bm{i}^\ast$, i.e., $\bm{f}^\ast = \bm{m}_{\bm{\theta}} (\bm{i}^\ast)$.
Therefore, the visual error to be considered in the standard \ac{vs} law~\eqref{eq:vs} is computed as 
\begin{equation}
    \bm{e} = \bm{f} - \bm{f}^\ast,
\end{equation}
whereas its norm, to be used in the large projection formulation~\eqref{eq:vs_norm}, is
\begin{equation}
    \eta = \Vert \bm{f} - \bm{f}^\ast \Vert
    \label{eq:norm_vis_error}
\end{equation}
where $\left\Vert \cdot \right\Vert$ denotes the Euclidean norm.

State-of-the-art systems, such as the YOLO algorithm~\cite{Redmon:cvpr:2016} considered in our work, can perform high-frequency and robust %
feature detection. 
However, such features are not truly informative of the real pose of the observed object and, thus, not enough for full servoing of the camera pose.
Typically, the bounding boxes detected by YOLO are rectangles centered on the image of the object of interest and aligned to the image borders.
Such bounding boxes do not include any information about the object orientation, as shown in Fig.~\ref{fig:yolo_orientation}.
An extended version of YOLO, YOLOv8~\cite{Jocher_Ultralytics_YOLO_2023}, computes bounding boxes that are oriented in the plane of the image. 
However, even in this case, the detected bounding box features do not contain information about the complete orientation of the detected object in all three rotation axes.
Thus, they are insufficient for controlling the full 6D pose of the camera and its motion in the Cartesian space.
One possible solution would be to refine the model $\bm{m}_{\bm{\theta}}$ by fine-tuning the parameters $\bm{\theta}$. 
However, this solution requires engineering work and the intervention of specialized scientists. 
Furthermore, the achievement of satisfactory results remains challenging.
Our solution, instead, proposes to learn how to overcome these limitations of the \revise{off-the-shelf \ac{dl} detection module} %
from given demonstrations of the desired task. 

\begin{figure*}[!t]
    \centering
    \includegraphics[width=0.8\textwidth]{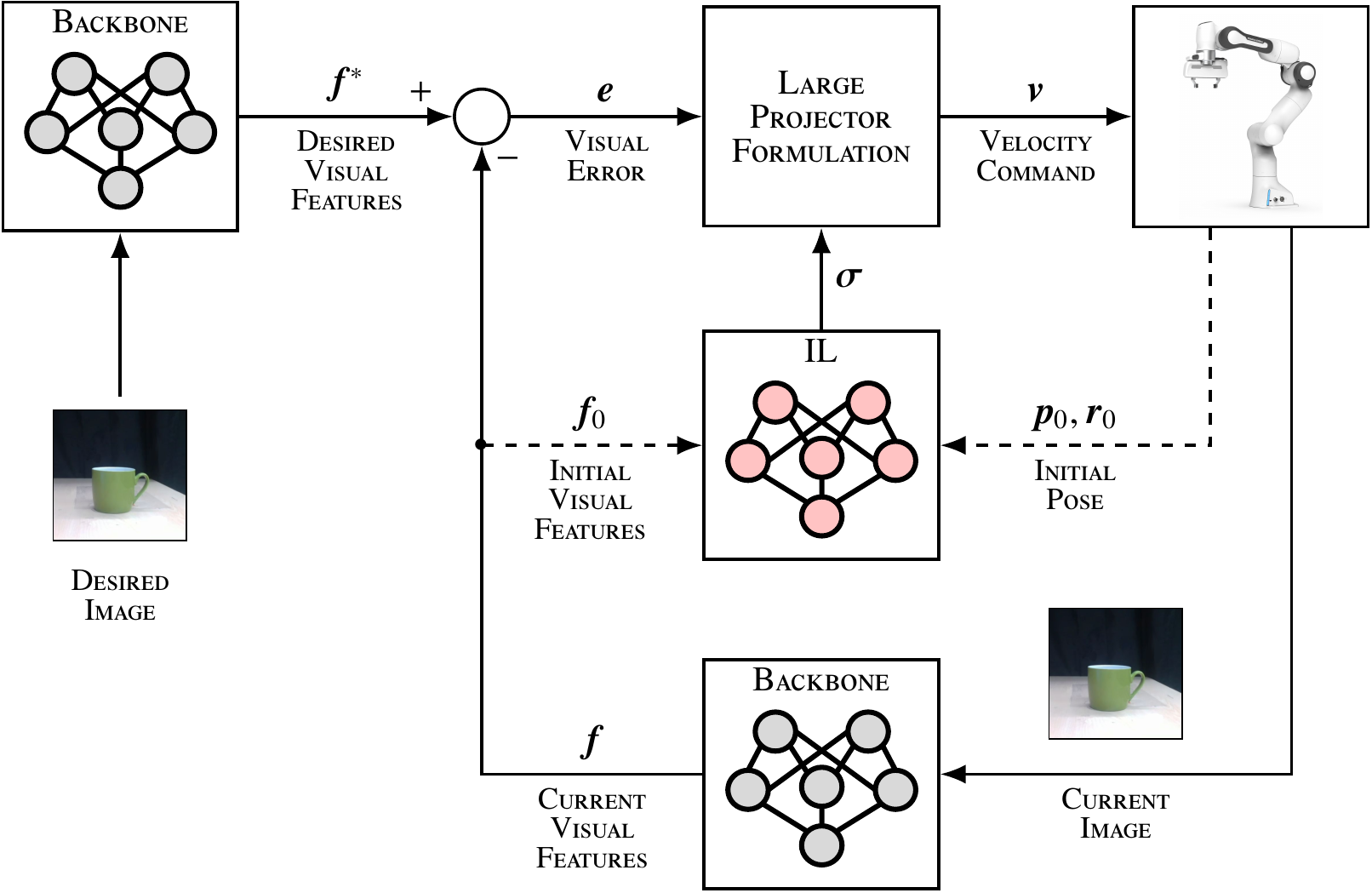}
    \caption{The proposed framework for \revise{\ac{ildvs}} exploits a \revise{detection model} (a frozen \ac{dl} network, implemented by YOLO) to extract features from raw images robustly and \ac{il} (implemented as a fine-tuned NODE network) to realize complex trajectories and overcome the limitation of the \revise{detection model}. The large projection formulation merges the output of the \revise{detection} and imitation strategy in a closed-loop control law resulting in accurate and converging robot movements.}
    \label{fig:approach}
\end{figure*}
\subsection{Overcoming the \revise{detection} limits through imitation}\label{sec:imitation}

To overcome the limitations of \ac{dl}-based %
detectors like YOLO, we leverage the information contained in a set of 
human-demonstrated
trajectories.
In particular, the corrective action for servoing the 3D orientation can be imitated from demonstrations of the full \ac{vs} behavior.
Such demonstrations are contained in a dataset with this shape:
\begin{equation}
    \mathcal{D} = \left\{ \bm{f}_{t}^{n}, \bm{p}_{t}^{n}, \bm{r}_{t}^{n} \right\}_{t,n}^{T,N},
    \label{eq:dataset}
\end{equation}
where the subscripts $t$ and $n$ denote the $t$-th sample of the $n$-th demonstration, respectively; $N$ is the total number of demonstrations and $T$ is the length (expressed as the total number of samples) of each demonstration; $\bm{f}$ indicates the visual features vector (as defined in Sec.~\ref{sec:backbone}), and $\bm{p}\in\mathbb{R}^3$ is the robot's end-effector's position. The end-effector orientation $\bm{r} \in \mathbb{R}^3$ is obtained by projecting a unit quaternion $\bm{q} \in\mathbb{S}^3$ into the \textit{tangent space} placed at the goal quaternion using the so-called \emph{logarithmic map}~\cite{ude2014orientation}. 
Such quantities are obtained by showing the full desired behavior to the robot by using, e.g., teleoperation or kinesthetic teaching. 
During the collection of demonstrations, the object of interest is placed at a fixed location w.r.t. the robot, and the object is always maintained in the camera field of view.

The \ac{il} strategy is realized by augmenting the \revise{detection} model with additional layers. 
More in detail, we augment the \revise{detector architecture} %
with a Neural Ordinary Differential Equation (NODE) solver~\cite{chen2018neural} used for \ac{il}. The additional NODE layers are trained on the dataset ${\cal D}$.
NODE has previously been used for IL~\cite{auddy2023continual} because it can be trained easily with only a few demonstrations, is extremely fast during inference, and exhibits accurate empirical performance for real-world full 6 degrees-of-freedom trajectory learning tasks~\cite{auddy2024scalable}.
The lack of mathematical stability guarantees of the trajectories predicted by a stand-alone NODE is addressed in our approach by the large projector formalism that guarantees that the position trajectory of the robot will not diverge, as discussed in Sec.~\ref{sec:ext_setup}.
Hence, we do not use other alternatives such as~\cite{urain2020imitationflow} that assure stability but have a much slower inference speed~\cite{auddy2023continual}.

NODE assumes that the training data are instances of a nonlinear dynamical system that maps a generic input state $\bm{x}$ into an output that consists of its time derivative $\dot{\bm{x}}$. 
In our setting, the state at time sample $t$ is $\bm{x}_t = \left(\bm{f}_t^{\top},\bm{p}_t^{\top},\bm{r}_t^{\top}\right)^{\top}$. %
To accurately approximate the underlying dynamics, NODE optimizes a set of parameters $\bm{\vartheta}$ by minimizing a sum-of-square-error loss. 
It is worth mentioning that, having projected unit quaternions in the tangent space, we can readily use the dataset $\mathcal{D}$ as in~\eqref{eq:dataset} to train NODE. 
This is a common strategy in robotics~\cite{ude2014orientation, huang2020toward, saveriano2019merging, Wang2022Learning}, which is also effective for NODE~\cite{auddy2023continual}. %
After training NODE, the rotation component is transformed back into unit quaternions using the so-called \emph{exponential map}~\cite{ude2014orientation, auddy2024scalable}. 
While training NODE, in each iteration we extract from the $N$ demonstrations in $\mathcal{D}$ a short contiguous segment of length $T_s$, %
obtained by drawing from $\mathcal{D}$ elements at random temporal locations $T_s$, $T_s \ll T$~\cite{auddy2023continual}.
We then concatenate each element of $\mathcal{D}_s$ into the vectors $\bm{x}_t^n = \left({\bm{f}_t^n}^{\top},{\bm{p}_t^n}^{\top},{\bm{r}_t^n}^{\top}\right)^{\top}$, $t=1,\ldots,T_s$, $n=1,\ldots,N$. %
Given the input vectors $\bm{x}_t^n, \forall t,n$, NODE uses its internal neural network $\bm{n}_\vartheta$ (called \emph{target network}) to produce derivatives of the input that are then numerically integrated to produce a predicted trajectory $\hat{\bm{x}}_t^n, \forall t,n$. %
For training NODE, we used the mean squared error loss $\mathcal{L}$, defined as: %
\begin{align}
\mathcal{L} &= \frac{1}{2} \sum_{n=1}^{N}\sum_{t=1}^{T_s} \Vert \bm{x}^{n}_t - \hat{\bm{x}}^{n}_t\Vert^2_2 \nonumber \\
&= \frac{1}{2} \sum_{n=1}^{N}\sum_{t=1}^{T_s} \Vert\bm{f}^{n}_{t} - \hat{\bm{f}}^{n}_{t}\Vert^2_2 + \Vert\bm{p}^{n}_{t} - \hat{\bm{p}}^{n}_{t}\Vert^2_2  + \Vert\bm{r}^{n}_{t} - \hat{\bm{r}}^{n}_{t}\Vert^2_2 \, . \label{eq:node_loss}
\end{align}
Ultimately, the trained NODE's target network is the following model
\begin{equation}
    \bm{\sigma} = \bm{n}_{\bm{\vartheta}}\left(\hat{\bm{f}}, \hat{\bm{p}},\hat{\bm{r}}\right)
    \label{eq:node_tgt_out}
\end{equation}
that is initialized with the initial state of the system comprised of the output $\bm{f}_0$ of the pre-trained detection model (i.e., \revise{YOLO}) %
and the initial robot's position $\bm{p}_0$ and tangent space orientation $\bm{r}_0$.
As output, it produces the robot velocity $\bm{\sigma}\in \mathbb{R}^6$, {which is the time derivative of $\bm{p}$ and $\bm{r}$}, imitating the complex trajectories demonstrated in the dataset.
In subsequent steps of the robot's motion, NODE evolves in an open-loop fashion its internal belief of the current state of $\hat{\bm{f}}$, $\hat{\bm{p}}$, and $\hat{\bm{r}}$, while keeping on predicting the velocity $\bm{\sigma}$.

\subsection{Merging DL and IL with the large projector}\label{sec:control}

The DL-based \revise{detection model} %
and the NODE target network are deployed within the robot control loop, as shown in the schematic of our approach (Fig.~\ref{fig:approach}). 
Given the desired and current image, the \revise{DL-based detector} %
extracts the visual features, as in~\eqref{eq:model_backbone}, which are then used to compute the visual error $\bm{e}$ and its norm $\eta$. 
These values are needed to compute the primary task in~\eqref{eq:vs_normal_and_norm}.
The lower priority (secondary) task considers the corrective velocity $\bm{\sigma}$ as regressed from the NODE target network~\eqref{eq:node_tgt_out}.

In our scheme, the pre-trained YOLO \revise{model} %
represents the \ac{dl}-based detector. %
The higher-priority task uses the current features (detected from the camera image) to adapt to changes in the object's location.
This feedback term also ensures convergence to the desired visual features $\bm{f}^*$ (i.e., $\bm{e} \rightarrow 0$ for $t \rightarrow +\infty$).
At the same time, the NODE network realizes an open-loop \ac{il} strategy that lets the robot execute more complex motions without affecting the convergence.

\section{Experimental setup}\label{sec:ext_setup}

In this section, we describe in detail our setup, including the hardware and software systems used in our experiments. 
We describe how we collect demonstrations, train NODE, and specify the metrics used for evaluation.

\subsection{Hardware and software components}

We use a Franka Emika Panda robot, a 7-degrees-of-freedom robotic arm. %
It features advanced force sensing and collision detection capabilities, making it safe and suitable for manipulation tasks in collaborative environments. 
The robot is fixed on a tabletop and equipped with an Intel RealSense D435 camera at the end effector. %
We run our image detection pipeline on a computer with an Intel i5-7640X CPU, 32GB RAM, and an NVIDIA GeForce RTX 4060 Ti GPU. 
The robot control software runs on a separate computer with a real-time OS kernel on the same network.

Our software is implemented using ROS noetic and we use the \texttt{franka\_example\_controllers}\footnote{\url{https://wiki.ros.org/franka_example_controllers}} package to communicate with the robot. The ROS interface accepts pose-level input, which we obtain by integrating the velocity command, computed as in~\eqref{eq:vs_norm}.
Furthermore, we experimentally verified that position and orientation tasks are decoupled in~\eqref{eq:vs_norm}. 
This is because the features from YOLO %
(obtained from the squared bounding box) do not contain information on the orientation; \revise{recall} Fig.~\ref{fig:yolo_orientation}. 
Therefore, as a difference from the {general} block diagram in Fig.~\ref{fig:approach}, the velocity output of NODE is split into its linear and angular parts; the first {actually} enters the large projection formulation, whereas the latter goes directly to the robot. 
This {implementation detail} implies that the control structure maintains the robot {stability} in position, while the execution of the orientation is delegated to the mere imitation strategy.
Nevertheless, {it is worth mentioning that} in our experiments, we observe that, with few demonstrations, the robot can perform safe behaviors even in orientation. %

We use the \texttt{realsense2\_camera}\footnote{\url{https://wiki.ros.org/realsense2_camera}} package to communicate with the camera and run YOLO with the \texttt{darknet\_ros}\footnote{\label{darknet}\url{http://wiki.ros.org/darknet_ros}} package, \revise{capturing RGB camera} images with a size of $640 \times 480$~pixels \revise{at} a framerate of $30$~Hz.  
\revise{
We use the standard calibration procedures provided by the \texttt{cv2}\footnote{\url{https://opencv.org/}} Python library to determine the camera's intrinsic parameters (consisting of the focal length and the central point).
Instead, to determine the extrinsic camera parameters (i.e., the pose of the camera with respect to the robot's gripper), we use the \texttt{aruco\_ros}\footnote{\url{https://wiki.ros.org/aruco_ros}} package and the \texttt{cv2} library.}
NODE is implemented and trained in PyTorch. 
\revise{Our open-source code, including the necessary software dependencies and calibration scripts, is available at \codeurl{}}.

\begin{figure}[!t]
\centering
\subfloat[Initial image of the mouse.]{\includegraphics[width=0.48\columnwidth]{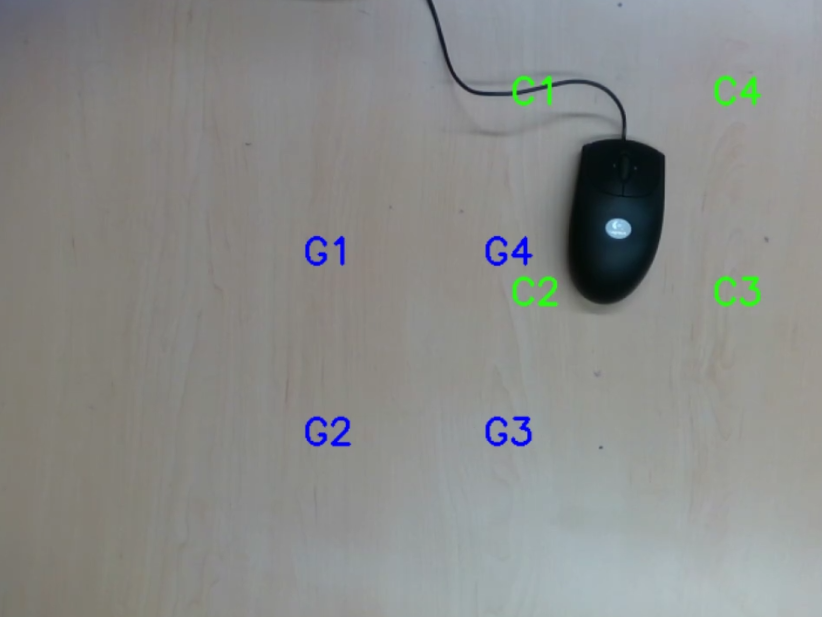}
\label{fig:demo_setup_mouse_init}}
\hfill
\subfloat[Final image of the mouse.]{\includegraphics[width=0.48\columnwidth]{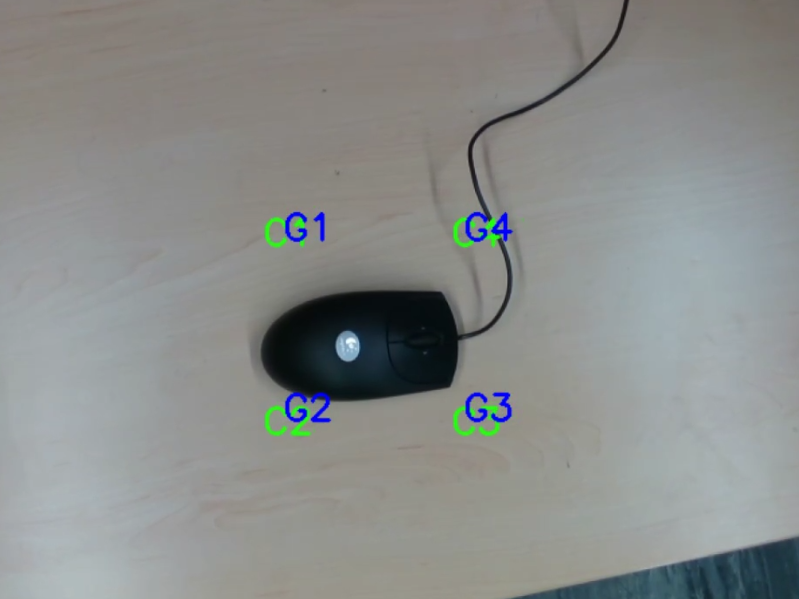}
\label{fig:demo_setup_mouse_final}}\\
\subfloat[Initial image of the cup.]{\includegraphics[width=0.48\columnwidth]{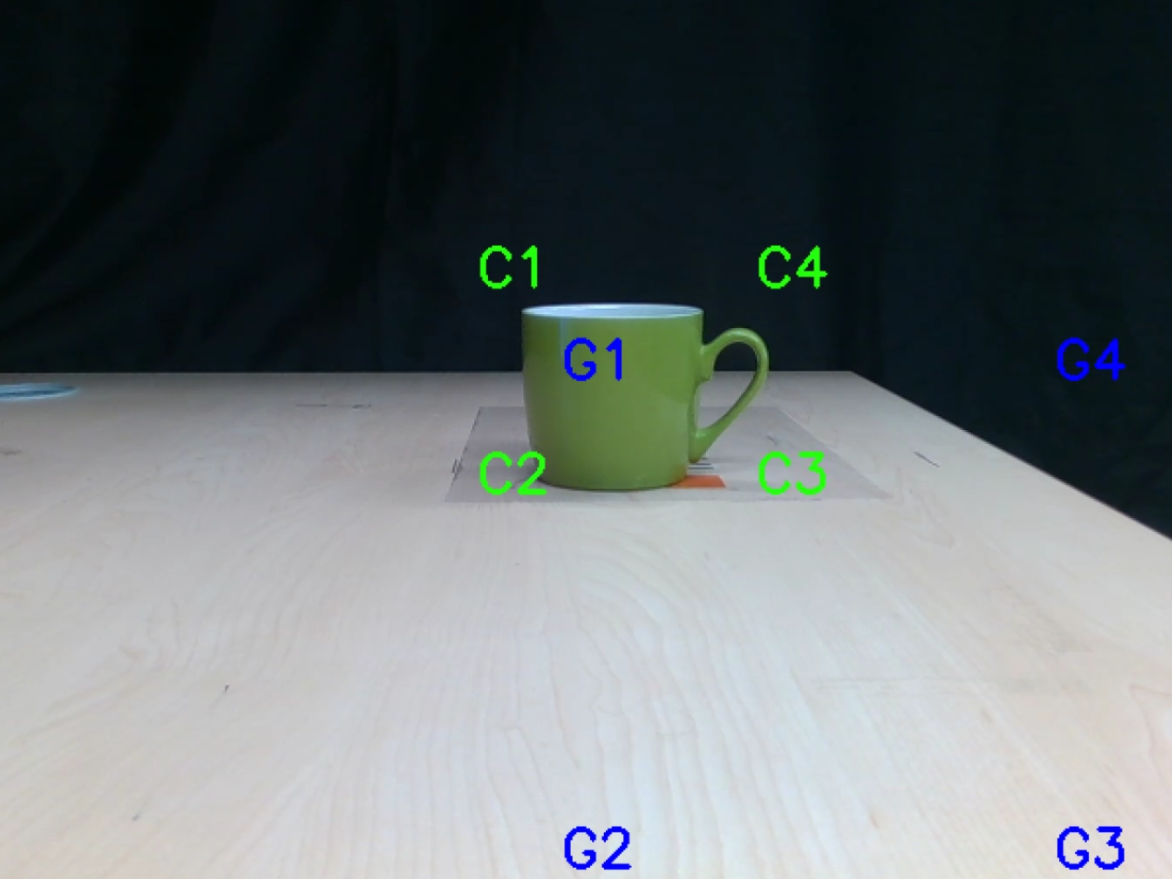}
\label{fig:demo_setup_cup_init}}
\hfill
\subfloat[Final image of the cup.]{\includegraphics[width=0.48\columnwidth]{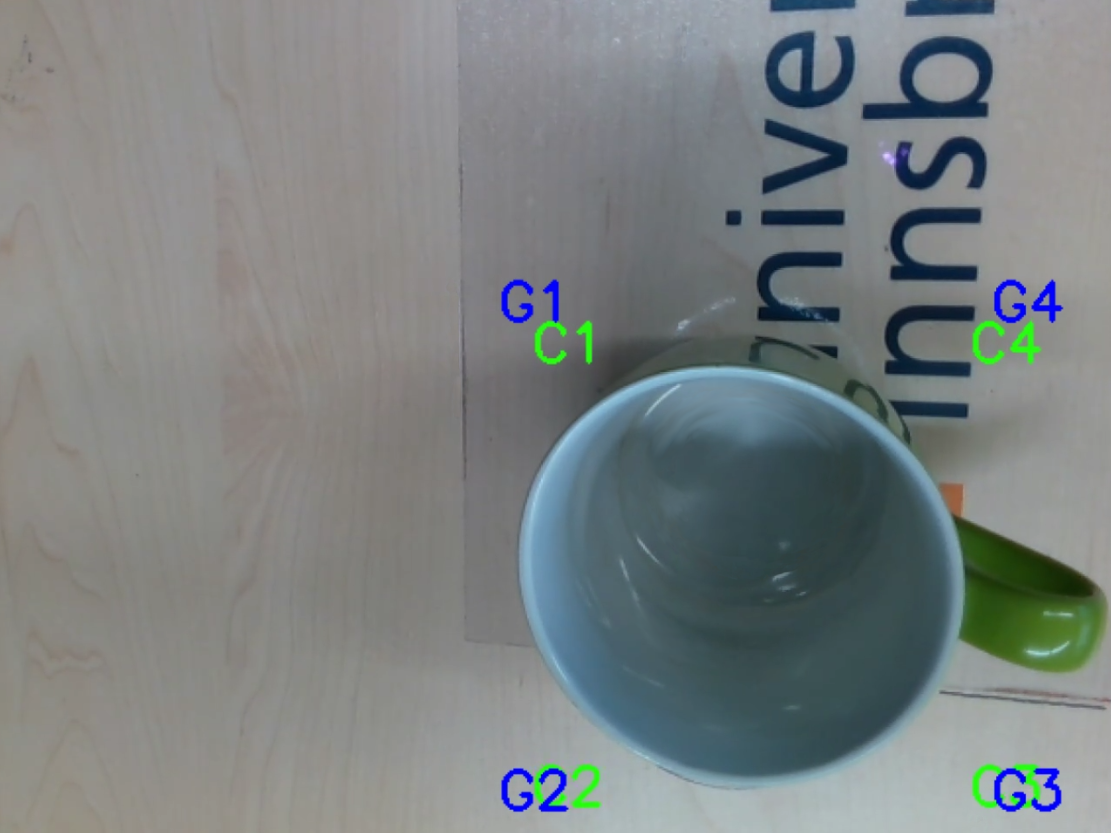}
\label{fig:demo_setup_cup_final}}
\caption[]{
Initial (left) and final images (right) captured by the robot camera in the experiments with the mouse (top) and the cup (bottom). Desired visual features are shown in blue and denoted with the letter ``G'', whereas the current visual features are the green letters ``C''. 
}
\label{fig:demo_setup}    
\end{figure}

\subsection{YOLO \revise{detector}}\label{subsec:yolo_backbone}

The YOLO \revise{detector} %
in our setup~(see Fig.~\ref{fig:approach}) uses the \texttt{yolov2-tiny}~\cite{jiang2022review} model pretrained on the COCO dataset~\cite{lin2014microsoft}. 
The underlying YOLO network can be easily changed to any of the other pre-trained models provided by the \texttt{darknet\_ros}\footref{darknet} package. 
In our experiments, 
we select the object of interest
(e.g., the mouse or the cup) from the list of all objects detected by YOLO and use the detected features of this object for VS.

\revise{
The rectangular bounding boxes originally predicted by YOLO often show significant variations in aspect ratio and vertex positions, even in consecutive frames that appear visually identical. 
These spurious changes cause abrupt jumps in the computed visual error that can disrupt closed-loop control dynamics and impede target convergence. 
This issue is particularly critical near the end of the robot's trajectory, where the current bounding box nearly overlaps with the target bounding box, potentially leading to convergence problems. 
Therefore, we convert the rectangular bounding boxes into squares with side lengths equal to the larger dimension of the original rectangle while maintaining the same center point.
To further mitigate noise in the detected visual features (i.e., vertices of the bounding box), we apply average filtering over the past $50$ frames. As the object of interest is not subject to rapid movements in the camera frame, smoothing the error signal facilitates convergence without compromising the robot's speed, as demonstrated in the supplementary video.
}

We use the vertices of the resulting bounding box as features in our VS scheme.
In the presentation of our results, such features are denoted with ``$\text{C}_i$'' whereas their desired counterparts are ``$\text{G}_i$'', with $i=1,\dots,4$.

\subsection{Collection of demonstrations} 

\begin{figure}[!t]
\centering
\includegraphics[width=0.9\columnwidth]{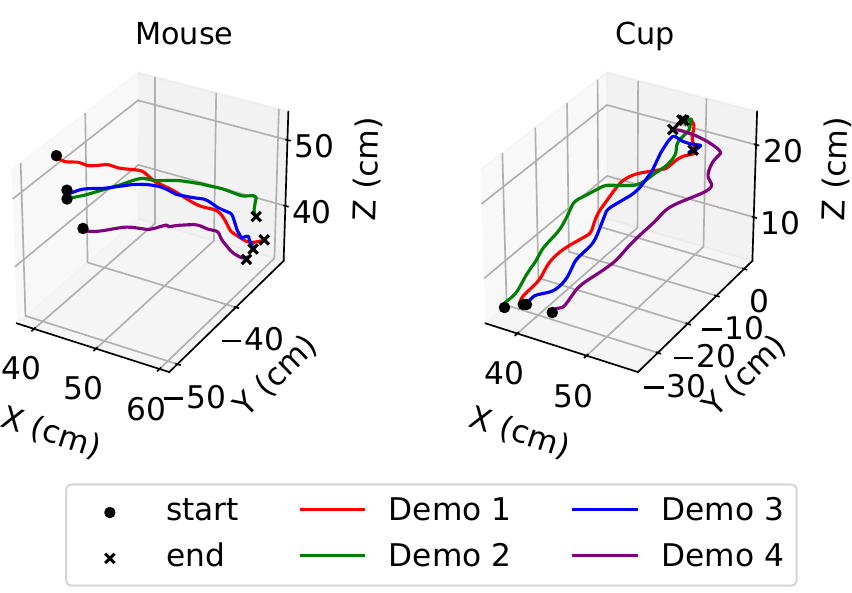}
\caption[]{\color{black}
Position trajectories of demonstrations provided for the ``Centering the mouse in the image''~(left) and ``Dropping an object in the cup''~(right) tasks. Diversity is introduced by starting from different initial poses and also through the differences between each kinesthetic demonstration.
}
\label{fig:demo_diversity}    
\end{figure}

For training NODE, we collect demonstrations via kinesthetic teaching~\cite{Billard:handbook:2008}. 
The object under consideration is placed in a specific location, and a human user physically guides the robot's end-effector from an initial pose to the desired final pose. 
We collect two sets of demonstrations corresponding to two different objects.
In the first set, a computer mouse is placed on the table with the robot's camera looking down at the mouse; kinesthetic demonstrations are provided so that the image of the mouse is rotated 90$^{\circ}$ clockwise in the final pose of the robot.
The latter set of demonstrations is collected using a cup. 
In the initial pose for these demonstrations, the camera looks side-on at a cup on the table; in the final one, the robot's end-effector is positioned on top of the cup with the camera looking down. 
See Fig.~\ref{fig:demo_setup} for a visual reference of the initial and final poses for the 
mouse and cup demonstrations.
Each set contains four demonstrations used to train a NODE (one for each set).
\revise{
Each demonstration consists of a sequence of $500$~steps.
In each set, the object's position remains unchanged but diversity is introduced into the collected data by starting each demonstration from a different pose.
Furthermore, manual kinesthetic demonstrations result in different trajectories and also introduce diversity into the training data.
The position trajectories of the demonstrations collected for the two tasks are depicted in Fig.~\ref{fig:demo_diversity}.
}

\begin{figure}[!t]
\centering
\includegraphics[width=0.65\columnwidth]{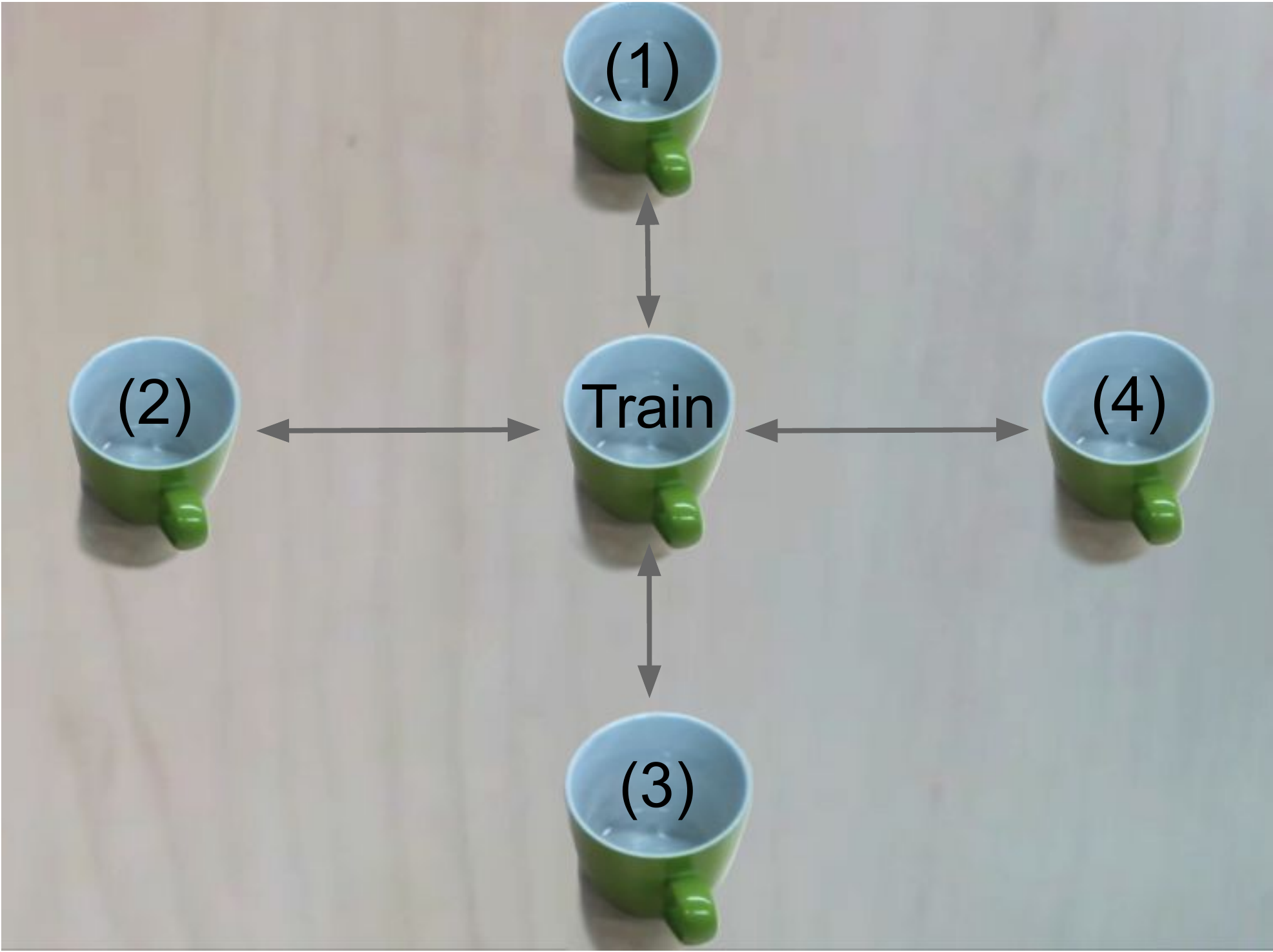}
\caption{The object position used during the collection of the demonstrations is the one at the center, whereas the four novel object positions used for the evaluation are $15$ cm off the center.}
\label{fig:cup_train_novel_positions}
\end{figure}

\subsection{NODE training}
The NODE target network predicts the derivatives of the inputs, and during training, numerical integration is used to generate trajectories from the predictions~\cite{auddy2023continual}. 
We use a NODE target network with 2 hidden layers containing 256 neurons each and ReLU activations. 
In each demonstration $n$ and each step $t$ of the training data, 
the inputs and outputs of the NODE are 10-dimensional, consisting of the upper left and lower right vertices of the bounding box
(i.e., the visual features) $\hat{\bm{f}}^n_t \in \mathbb{R}^4$ in image coordinates normalized to lie within $[0.0,100.0]$, the position (in cm) of the robot's end-effector in the task space $\bm{p}^n_t \in \mathbb{R}^3$, and the rotation vector $\bm{r}^n_t \in \mathbb{R}^3$ obtained by projecting the orientation quaternions of the end-effector to the local tangent space, as described in Sec. \ref{sec:imitation}. 
Following~\cite{auddy2023continual}, we scale the rotation vectors by a constant factor of 100.0 so that all input features are of comparable magnitudes. 
For each recorded demonstration set (mouse and cup), we train a NODE for $2\times 10^4$ iterations with a learning rate of $5 \times 10^{-4}$ using the loss defined in \eqref{eq:node_loss}.

Note that the sides of the bounding boxes detected by YOLO are always parallel to the image sides (the image coordinates of only the upper left and lower right vertices of the bounding boxes are predicted). Consequently, the visual features that are recorded to train the NODE are also 4-dimensional.
Our VS scheme uses a general representation of a bounding box consisting of the features corresponding to all 4 vertices. Therefore, during inference, we compute the 8-dimensional visual features by deriving the coordinates of the upper right and lower left vertices from the YOLO predictions.

\subsection{Evaluation protocol}\label{sec:ext_setup_eval}

Our analysis compares the performance of three \ac{vs} schemes.
The first is denoted with \ac{iil} and uses a NODE instance trained on the demonstrations to control the robot in an end-to-end fashion.
The second one is a classic \ac{vs} scheme where YOLO provides the required visual feedback; following the literature~\cite{Collowet:tro:2011, Bateux:ral:2017}, we call it \ac{dvs}, as it is a \emph{direct} approach considering the whole image as input.
Finally, our proposed method, augmenting the \ac{ilvs} scheme~\cite{Paolillo:icra:2022} with \ac{dl}-based direct measurement, is called \ac{ildvs}. \revise{It is worth mentioning that, for a fair comparison, all the approaches are fed with the square and filtered bounding boxes computed as discussed in Sec.~\ref{subsec:yolo_backbone}.}

We conduct separate experiments for the mouse and the cup.
For each experiment, we evaluate the performance of the three schemes for five different object positions: one as in the demonstrations, and four unseen positions, as shown in Fig.~\ref{fig:cup_train_novel_positions}.
We run each test for $T=700$ time steps, where $T$ is the demonstration length,  and measure the norm of the %
visual error. %
and the end-effector position and orientation error at the final step of the experiment.

\begin{figure*}[t]
\centering
\includegraphics[scale=0.45]{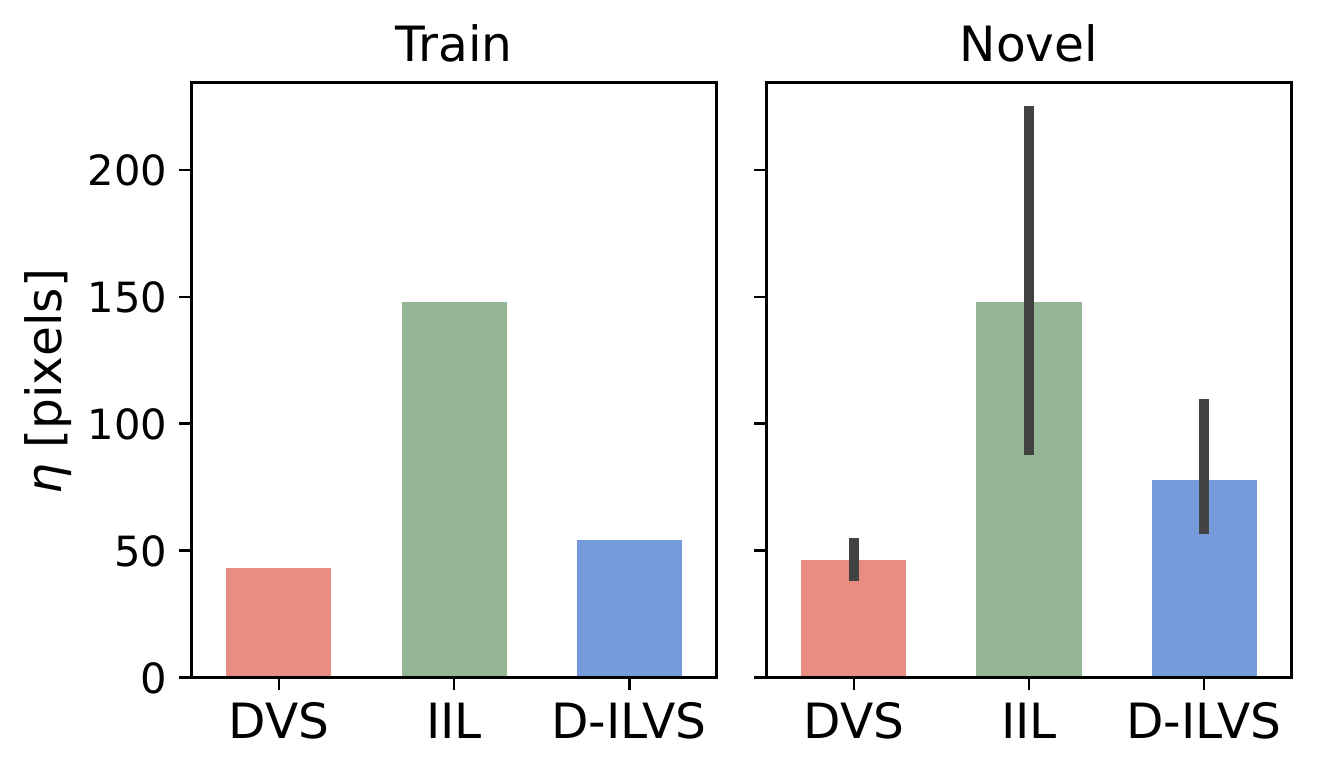}
\quad
\includegraphics[scale=0.44]{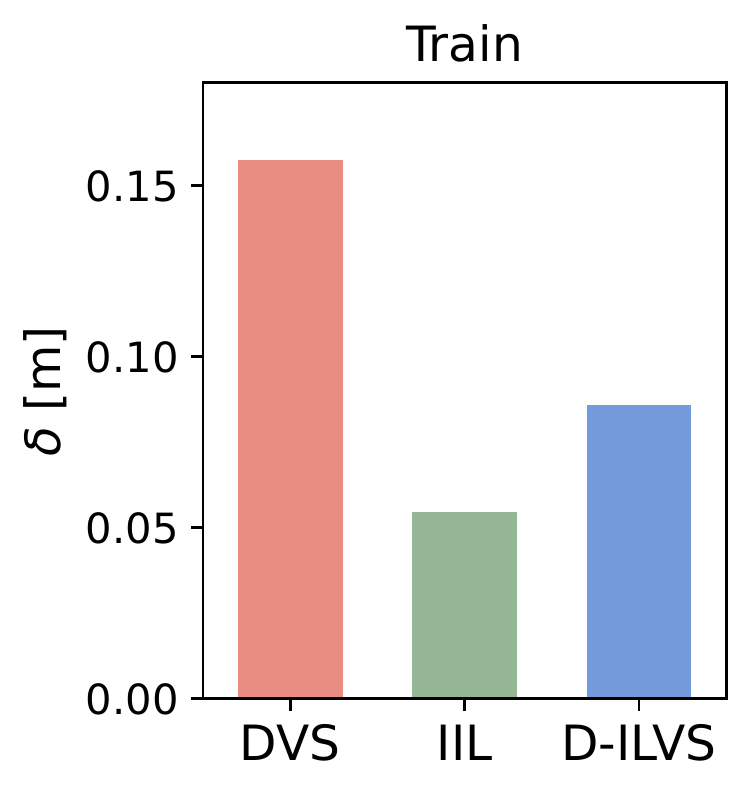}
\quad
\includegraphics[scale=0.45]{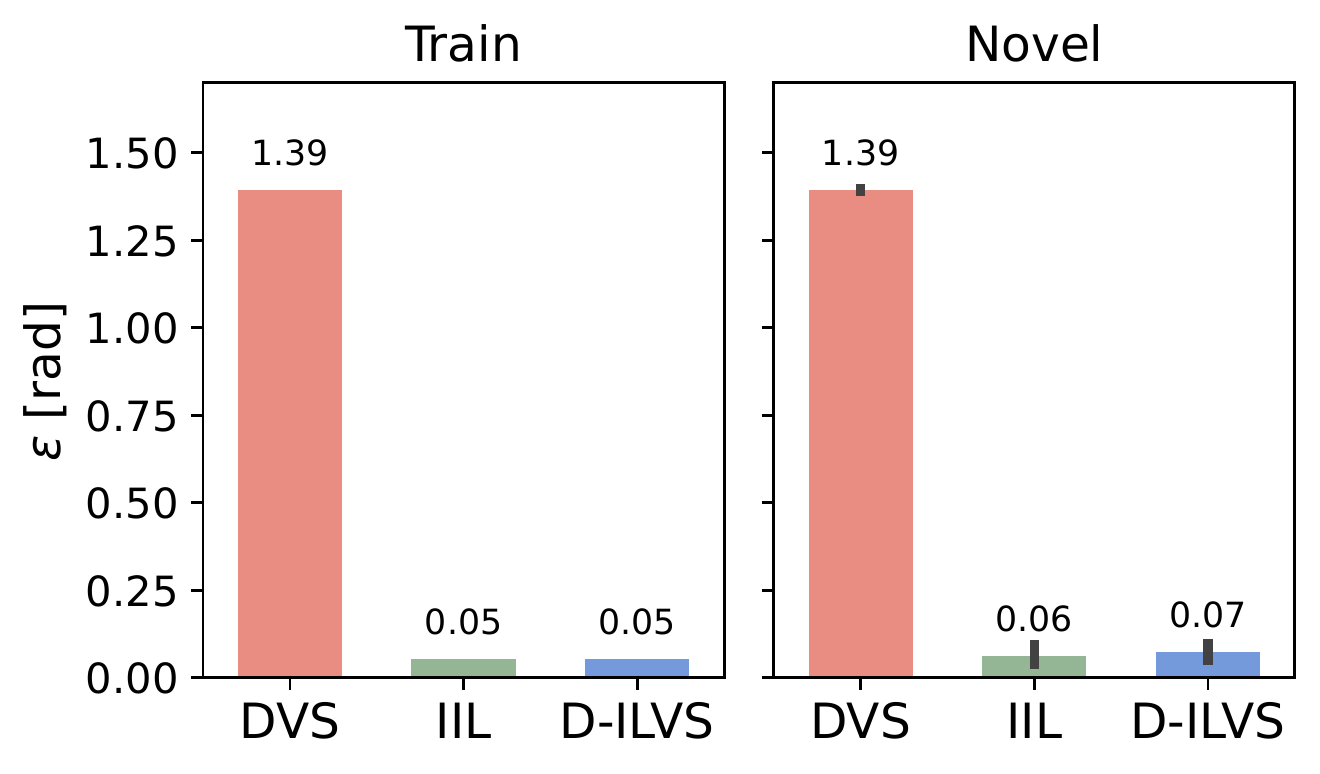}\caption[]{Centering the mouse in the image: the average norm of the final visual error (left), position (center), and orientation error (right) achieved with the three schemes, starting from similar trained positions or novel ones. 
\revise{All errors are computed at the final step of the robot's trajectory.}
Note that the end-effector position error cannot be computed for the novel positions.
{Colored boxes show the means and error bars show the 95\% confidence interval.}
}
\label{fig:mouse_bars}    
\end{figure*}
\begin{figure*}[ht!]
\centering
\includegraphics[scale=0.45]{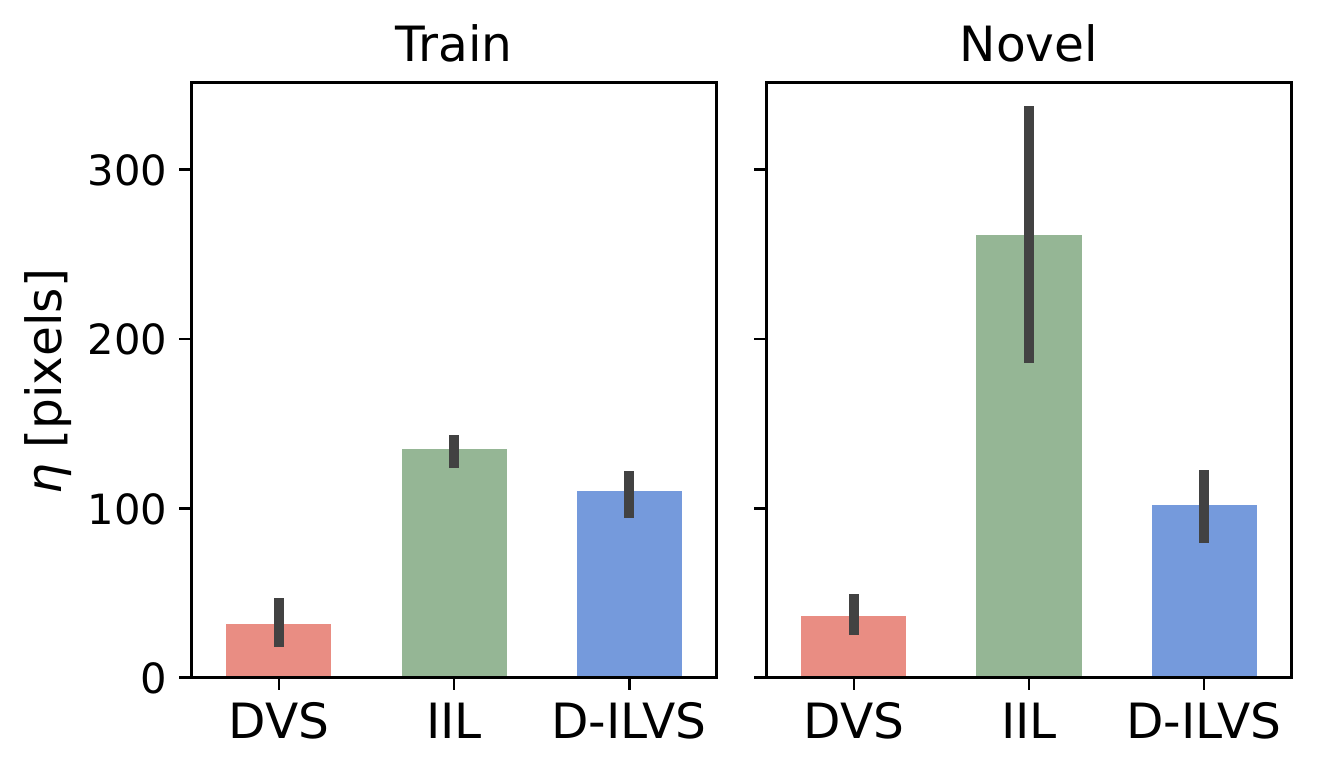}
\quad
\includegraphics[scale=0.44]{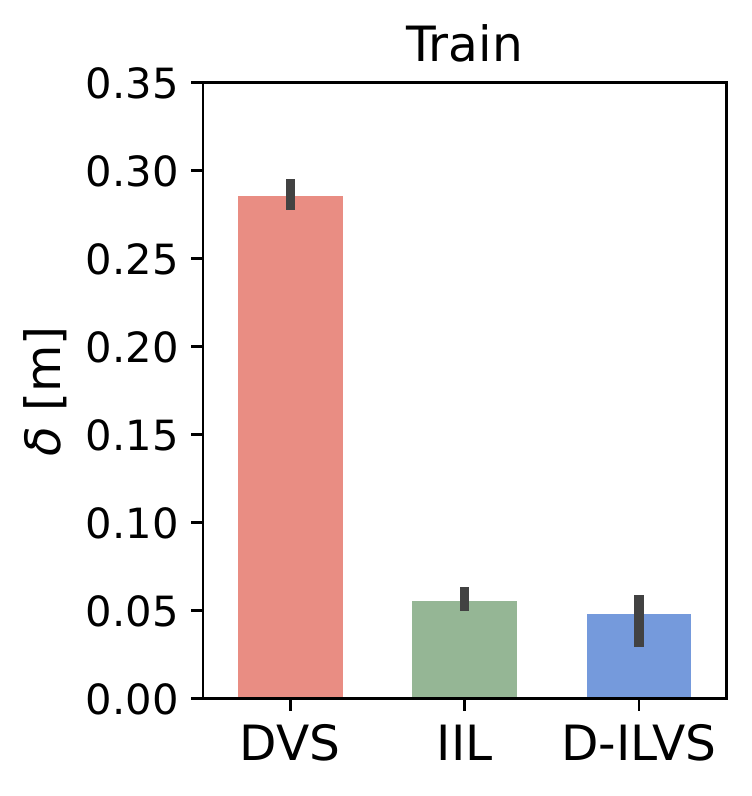}
\quad
\includegraphics[scale=0.45]{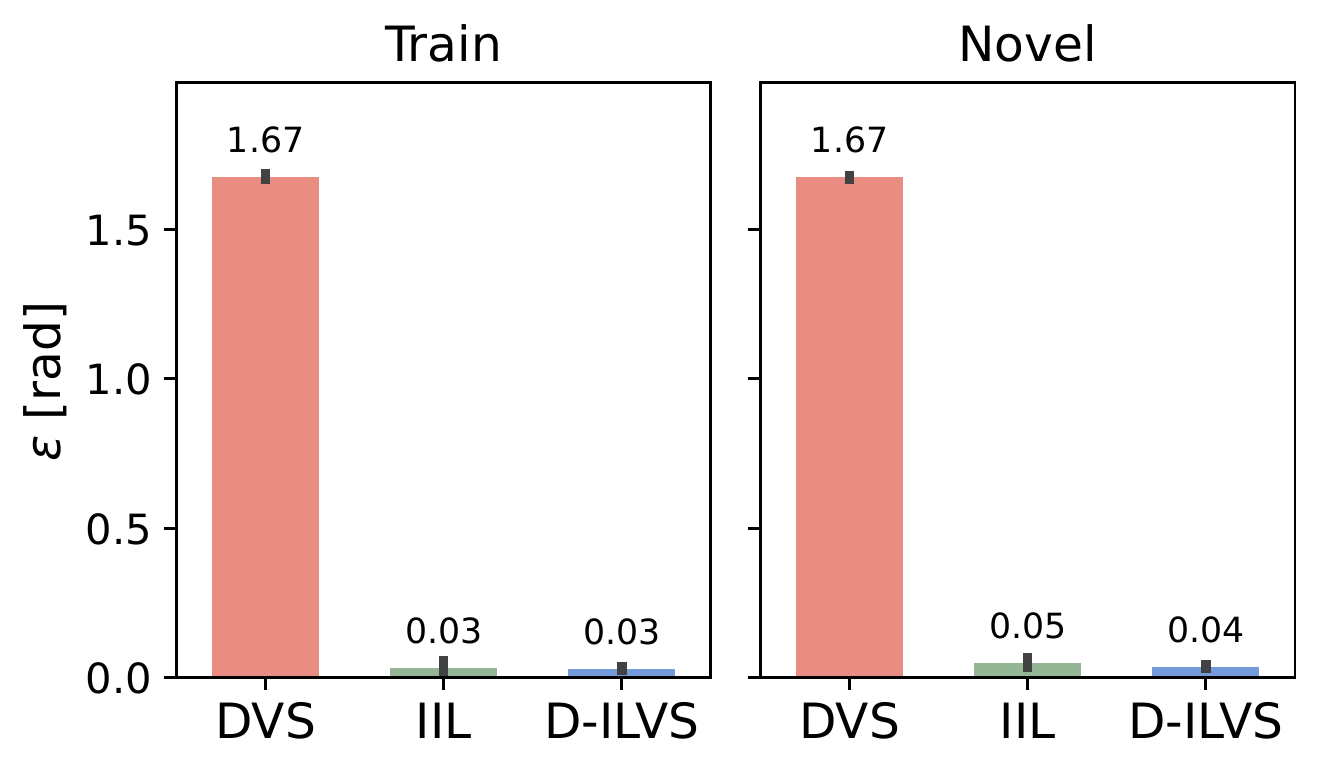}
\caption[]{Dropping an object in the cup: the average norm of the final visual error (left), position (center), and orientation error (right) achieved with the three schemes, starting from similar trained positions or novel ones. 
\revise{All errors are computed at the final step of the robot's trajectory.}
Note that the end-effector position error cannot be computed for the novel positions. Colored boxes show the means and error bars show the 95\% confidence interval.
}
\label{fig:cup_bars_0_others}    
\end{figure*}
\revise{All errors are computed at the final step of the robot's trajectory. In particular, the visual error norm $\eta$ is computed using~\eqref{eq:norm_vis_error}}.
The end-effector position error is computed as
\begin{equation}
    \delta = \Vert \bm{p}_T - \bm{p}_g \Vert_2
    \label{eq:position_error}
\end{equation}
where $\bm{p}_g \in \mathbb{R}^3$ is the Cartesian position of the robot at the last step of the demonstration (i.e., the ground truth desired position) and $\bm{p}_T \in \mathbb{R}^3$ is the final position reached during the evaluation.
The orientation error is computed using the orientation $\bm{q}_T \in \mathbb{S}^3$ of the robot in the last step of the experiment and $\bm{q}_\text{g} \in \mathbb{S}^3$ the orientation at the end of the demonstration (i.e., the ground truth desired orientation). 
The measure of the orientation error is computed as
\begin{equation}
    \varepsilon = %
    \left\Vert\ \log \left(\bm{q}_T \otimes \bar{\bm{q}}_\text{g}\right)\right\Vert_2
    \label{eq:ori_err}
\end{equation}
where $\log(\cdot)$ denotes the logarithmic map~\cite{ude2014orientation}, %
$\bar{\bm{q}}_g$ denotes the conjugation of quaternion $\bm{q}_g$, whereas the symbol `$\otimes$' is the quaternion product operator.
Note that the position error $\delta$ is not computed for novel object positions as there is no ground truth end-effector position $\bm{p}_g$ for these object positions.
However, the relative orientation of the robot's end-effector with respect to the object at the end of execution is expected to be the same irrespective of whether the object is placed at the demonstrated or novel locations. This enables us to measure the orientation error $\varepsilon$ for trained as well as novel object positions.

Furthermore, for the cup experiment, we measure the success of dropping a small object into the cup at the end of the robot's motion. 
Since resetting the robot after every evaluation may introduce some stochasticity, we evaluate each method on each object position three times. 
Finally, we also perform a qualitative evaluation of our \ac{ildvs} scheme in a cluttered scene where an object is to be dropped into a cup at different positions among several other objects.

\section{Experimental results}\label{sec:results}

During the evaluation of the different approaches described in Sec.~\ref{sec:ext_setup_eval}, the same trained NODE is used in our \ac{ildvs} approach as well as in \ac{iil} where NODE controls the robot in an end-to-end way. 
This section presents the results of the different approaches in the mouse and cup experiments.
Examples of the presented experiments can be viewed in the attached supplementary video. 

\subsection{Centering the mouse in the image}\label{sec:results_mouse}

\begin{figure}[!b]
\centering
\subfloat[Initial image.]{\includegraphics[width=0.475\columnwidth]{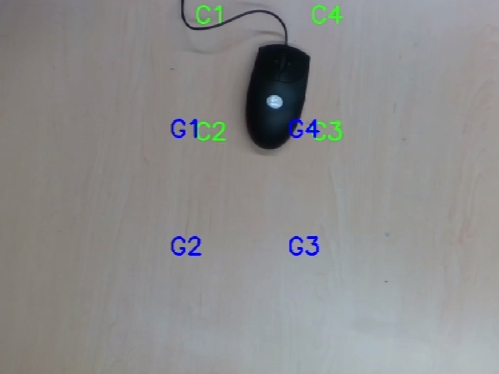}
\label{fig:mouse_qual_init}
}
~
\subfloat[Final image with DVS.]{\includegraphics[width=0.476\columnwidth]{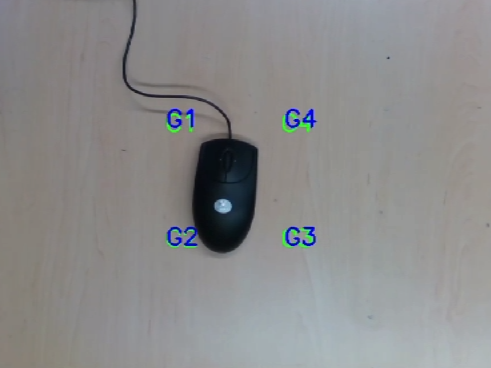}
\label{fig:mouse_qual_final_yolo}
}
\\
\subfloat[Final image with IIL.]{\includegraphics[width=0.475\columnwidth]{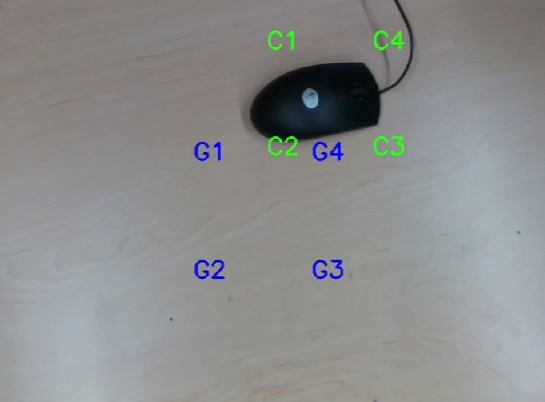}
\label{fig:mouse_qual_final_node}
}
~
\subfloat[Final image with \revise{\ac{ildvs}}.]{\includegraphics[width=0.475\columnwidth]{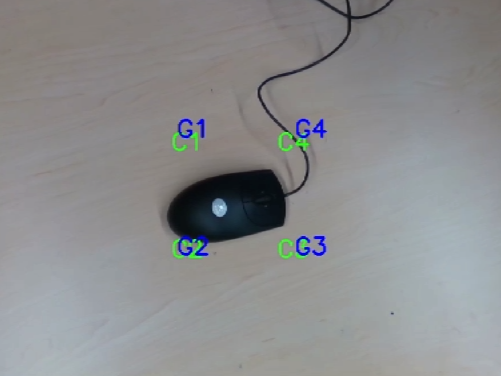}
\label{fig:mouse_qual_final_node_yolo}
}
\caption[]{
Centering the mouse in the image with novel mouse locations: initial and final images reached with the different methods.
}
\label{fig:mouse_qual}    
\end{figure}

The goal of the first set of experiments is to move the robot such that the image of the mouse (whose visual features are shown in green in Fig.~\ref{fig:mouse_qual}) coincides with its desired image (corresponding to a desired set of features, shown in blue in Fig.~\ref{fig:mouse_qual}).
Note that the task also requests the robot camera (and end-effector) to rotate in order to reach the desired relative camera-mouse pose. 
We evaluate each of the three schemes (DVS, IIL, \ac{ildvs}) on five different object positions, as described in Sec.~\ref{sec:ext_setup_eval}, and report the overall results in Fig.~\ref{fig:mouse_bars}. 
\ac{dvs} achieves low visual errors but fails to control the orientation correctly. 
\ac{iil} achieves low orientation errors but cannot adapt to novel object positions leading to a high visual error. 
In contrast, our \ac{ildvs} approach exhibits a low visual error, can adapt to novel object positions, and controls the robot orientation properly.

More in detail, \ac{dvs} achieves the lowest visual error as VS with YOLO alone can position the robot camera such that the current and desired visual features match very closely. 
However, as the features detected by YOLO do not have any information about the real orientation of the mouse, 
the yaw orientation of the robot's end-effector does not change at all. 
As a result, the final orientation reached in this experiment is very different from the desired one (compare the desired pose in Fig.~\ref{fig:demo_setup} with the final pose achieved by \ac{dvs} in Fig.~\ref{fig:mouse_qual_final_yolo}). 
The \ac{iil} approach can achieve low orientation error as the underlying NODE has been trained to achieve the demonstrated orientation. 
However, it cannot compensate for the changes in the novel mouse positions outside the demonstrations, resulting in high visual error.
This effect can also be qualitatively observed in Fig.~\ref{fig:mouse_qual_final_node}.
Our \ac{ildvs} approach can adapt to novel positions of the mouse.
At the same time, it utilizes the trained NODE to achieve the correct orientation as shown by the kinesthetic demonstrations.
This enables \ac{ildvs} to take advantage of both DVS and IL and achieve low visual and orientation errors, making it the best approach among the ones we evaluate.
Fig.~\ref{fig:mouse_qual_final_node_yolo} shows that \ac{ildvs} achieves a close fit to the desired visual features and the desired orientation.

\subsection{Dropping an object in the cup}\label{sec:results_cup}

\begin{figure}[b!]
\centering
\subfloat[]{\includegraphics[width=0.48\columnwidth]{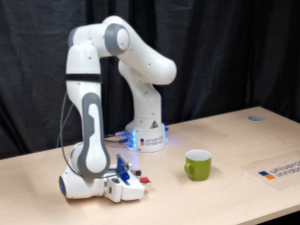}
\label{fig:qual_cup_1_yolo_robot}
}
\hfill
\subfloat[]{\includegraphics[width=0.48\columnwidth]{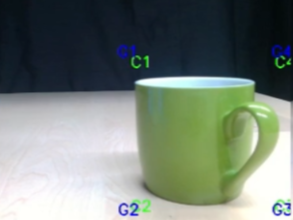}
\label{fig:qual_cup_1_yolo_internal}
}\\
\subfloat[]{\includegraphics[width=0.48\columnwidth]{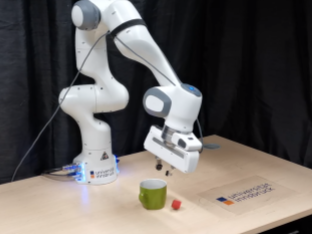}
\label{fig:qual_cup_2_node_robot}
}
\hfill
\subfloat[]{\includegraphics[width=0.48\columnwidth]{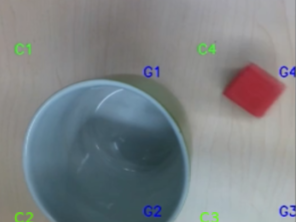}
\label{fig:qual_cup_2_node_internal}
}
\\
\subfloat[]{\includegraphics[width=0.48\columnwidth]{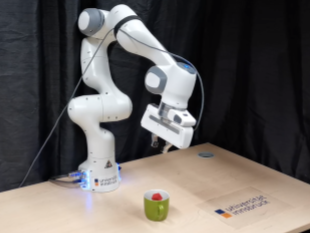} 
\label{fig:qual_cup_3_node_yolo_robot}
}
\hfill
\subfloat[]{\includegraphics[width=0.48\columnwidth]{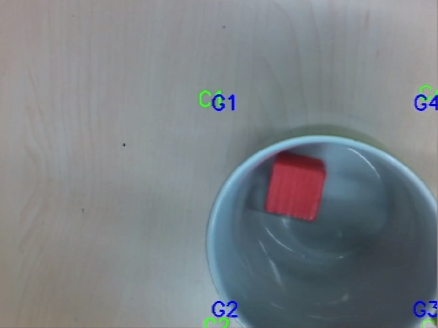}
\label{fig:qual_cup_3_node_yolo_internal}
}
\caption[]{
Dropping an object in the cup placed in a novel position: robot external view (left column) and the corresponding camera view (right column) as executed by \ac{dvs} (a,b), \ac{iil} (c,d) and \ac{ildvs} (e,f).
}
\label{fig:cup_qual}    
\end{figure}

\begin{figure}[!b]
    \centering    \includegraphics[width=0.48\textwidth]{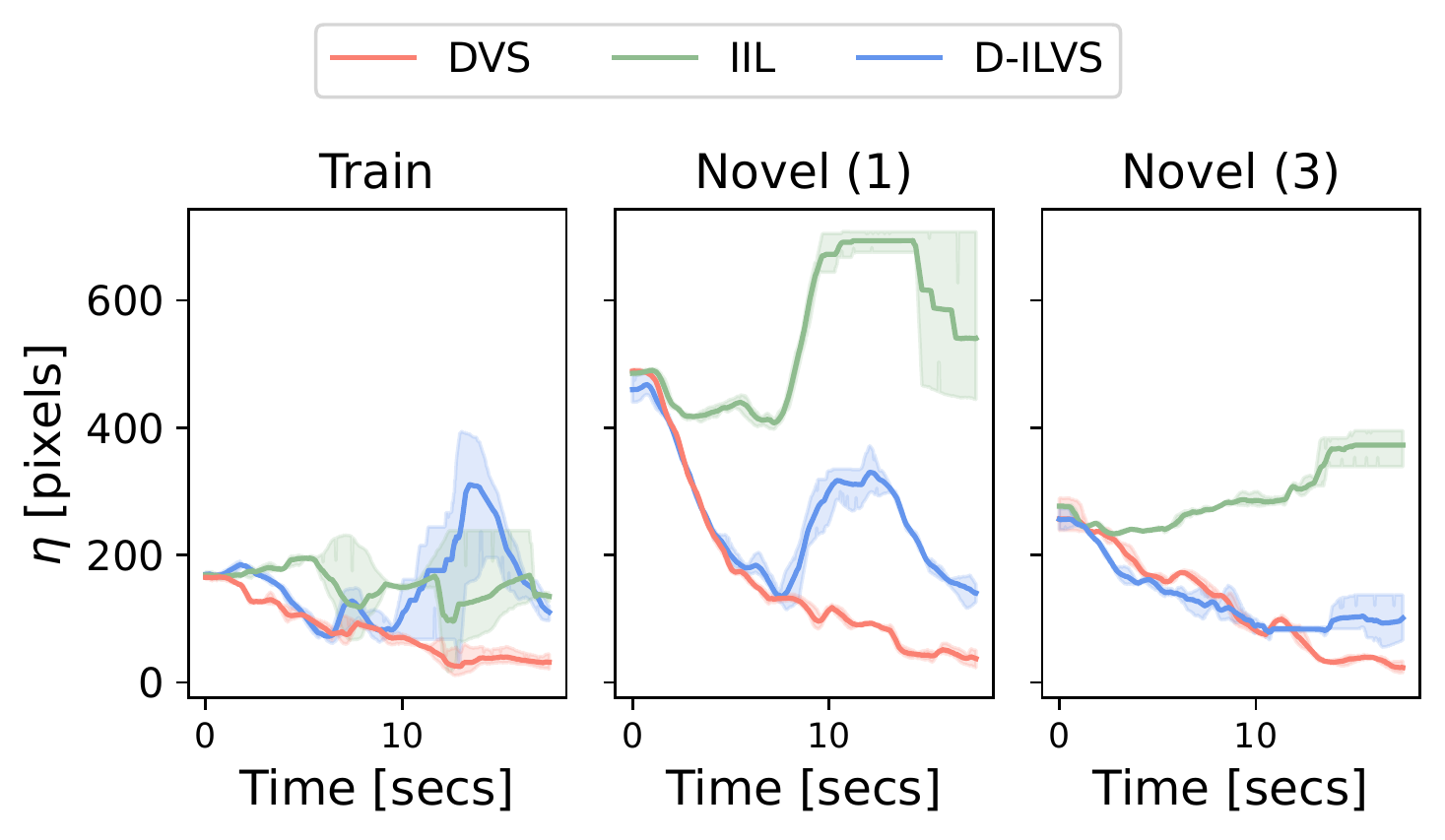}
    \caption{Drop an object in the cup: the visual error with the trained object position (left) and two novel positions (center and right).
    }
    \label{fig:cup_line}
\end{figure}

The second set of experiments aims to drive the robot end-effector on top of the cup and drop an object in it, leveraging the demonstrations recorded with the cup.
Note that these experiments present the additional challenge of performing nontrivial (e.g., nonlinear) trajectories.
The camera's initial and desired views are representative of two completely different camera-cup relative poses, as shown in Fig.~\ref{fig:demo_setup}.

\begin{figure*}[b!]
\centering%
\includegraphics[width=0.32\textwidth]{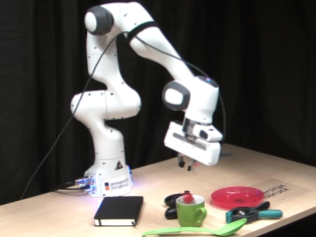}
~
\includegraphics[width=0.32\textwidth]{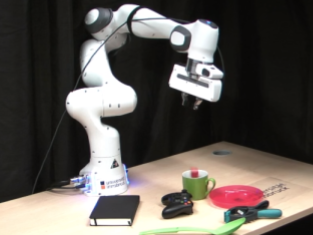}
~
\includegraphics[width=0.32\textwidth]{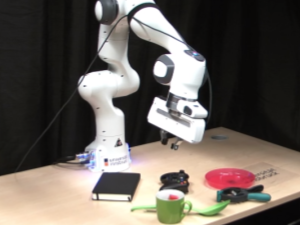}
\caption{With \revise{\ac{ildvs}}, the robot successfully drops the object into the cup placed in different positions of a cluttered table.
}
\label{fig:cup_cluttered}    
\end{figure*}

The quantitative evaluation is presented in Fig.~\ref{fig:cup_bars_0_others}. 
The different approaches are again evaluated for one trained object position and four novel object positions outside the demonstrations, as described in Sec.~\ref{sec:ext_setup_eval}.
\ac{dvs} achieves a low visual error but fails to move the end-effector above the cup and orient it properly, resulting in high position and orientation errors.
\ac{iil} achieves a low orientation error but cannot adapt to novel object positions leading to a high visual error. 
As expected, the visual error is much higher for the novel object positions; the orientation error remains low for both trained and novel object positions.
\ac{ildvs}, i.e., our approach, achieves low errors for both position and orientation (corresponding to limited visual errors) as it can adapt to novel object positions and orient the robot properly.

A qualitative evaluation is shown in Fig.~\ref{fig:cup_qual} and confirms the quantitative results.
The \ac{dvs} approach can align the visual features to their desired counterpart (Fig.~\ref{fig:qual_cup_1_yolo_internal}), but the robot ends up in a completely wrong pose (Fig.~\ref{fig:qual_cup_1_yolo_robot}).
As a result, it cannot drop the grasped object into the cup.
\ac{iil} attempts to realize the complex trajectory required to execute the dropping task, but it has poor accuracy (Fig.~\ref{fig:qual_cup_2_node_robot}).
In particular, for a novel object position, as the one shown in the figure, 
the robot is unable to adapt and drops the object outside the cup (Fig.~\ref{fig:qual_cup_2_node_internal}).
\ac{ildvs} (ours) shows the best performance among all the approaches, 
as it can cope with changing positions of the cup and drops the object successfully, as shown in Figs.~\ref{fig:qual_cup_3_node_yolo_robot} and~\ref{fig:qual_cup_3_node_yolo_internal}. 

Figure~\ref{fig:cup_line} shows the time evolution of the visual error for the trained object position and two novel object positions during a complete experiment 
(see Fig.~\ref{fig:cup_train_novel_positions} for a description of the trained and novel object positions).
\ac{ildvs}~(ours) achieves a low visual error for both novel as well as trained object positions, whereas \ac{iil} achieves much higher visual errors for novel object positions due to its inability to adapt. 
As expected, the visual error made by \ac{dvs} stays low throughout.

We also report the success rates of dropping the object into the cup for all methods (see Tab.~\ref{tab:drop_in_cup_new}). 
Each approach is evaluated 15 times as described in Sec.~\ref{sec:ext_setup_eval} (5~object positions for each of the 3~trials per object position) and we report mean values for success. 
A trial is considered $100\%$ successful if the object drops cleanly into the cup, $50\%$ successful if the object hits the cup's rim but eventually falls inside, and $0\%$ otherwise. 
Tab.~\ref{tab:drop_in_cup_new} shows that our \ac{ildvs} approach achieves near-perfect results, while \ac{iil} achieves a much lower score since it is unsuccessful in dropping the object into the cup placed at novel positions; \ac{dvs} is never able to drop the object into the cup and gets a score of $0$.
\begin{table}[t]
\centering
\caption{Success rates for dropping an object into the cup.}
\label{tab:drop_in_cup_new}
\begin{tabular}{lccc}
\toprule
\multirow{2}{*}{\sc{Approach}} & \multicolumn{3}{c}{\sc{Success Rate [\%]}} \\ \cmidrule(lr){2-4}
& \sc{Train} & \sc{Novel} & \sc{Overall}\\
\midrule
\ac{dvs} & 0.00 & 0.00 & 0.00 \\
\ac{iil} & 83.33 & 0.00 & 16.67 \\
\ac{ildvs} & \bfseries 100.00 & \bfseries 95.83 & \bfseries 96.67 \\
\bottomrule
\end{tabular}
\end{table}

\begin{figure}[b!]
\centering
\includegraphics[width=0.48\columnwidth]{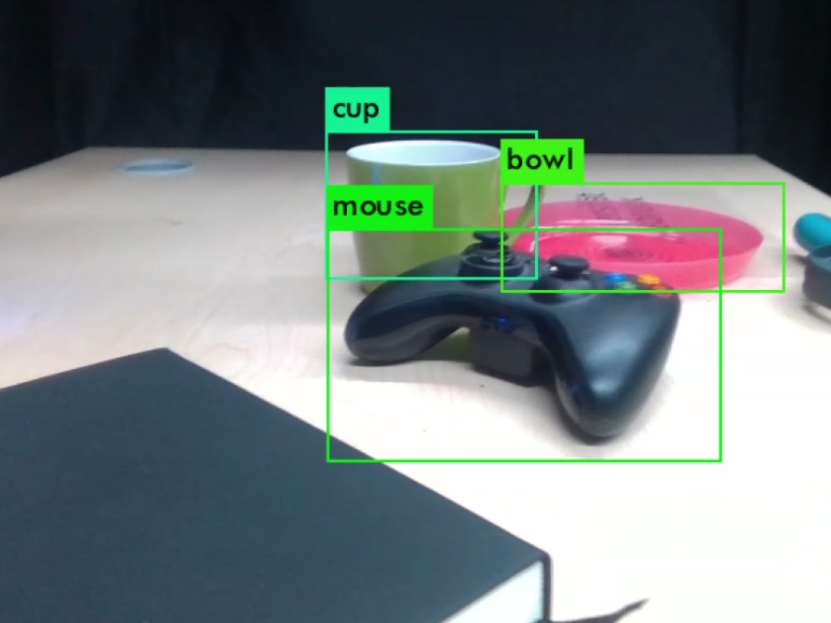}
\hfill
\includegraphics[width=0.48\columnwidth]{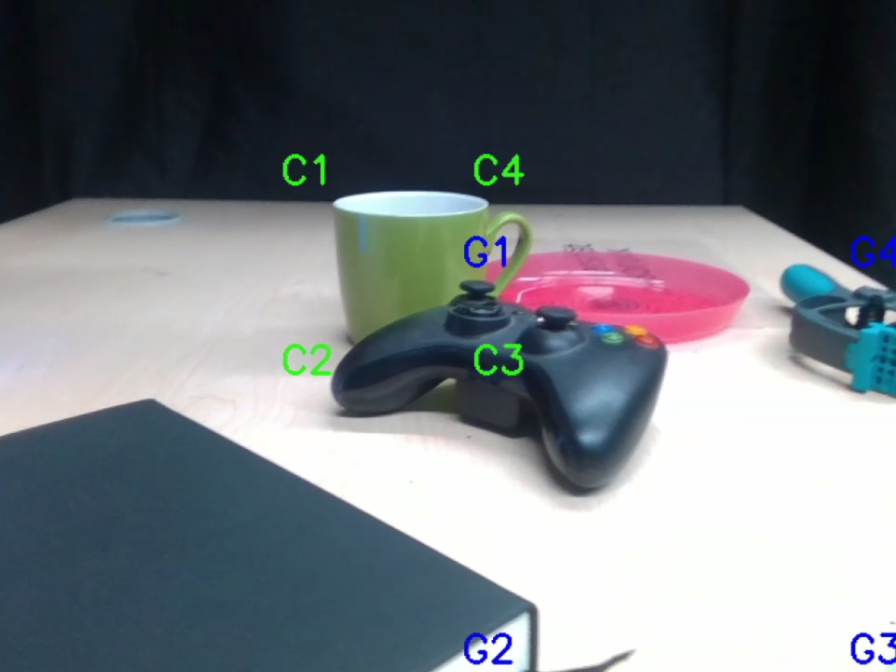}
\caption[]{YOLO detects a cup robustly even in a cluttered scene ~(left), providing the required visual features shown with green letters `C'~(right); the desired features are also shown, with blue letters `G'.
}
\label{fig:cup_clutter_internal}   
\end{figure}

\subsection{Handling cluttered scenes}

We execute the experiments presented in the previous section in a cluttered setting and qualitatively evaluate the effectiveness of our \ac{ildvs} approach. 
The cup is placed on the table among several other objects, such as a book, a plate, a clamp, a spatula, and a game controller. 
In different trials, the location of the cup is varied among the other objects (see Fig.~\ref{fig:cup_cluttered}). 

The pre-trained YOLO object detection model identifies multiple objects in the scene, as shown in Fig.~\ref{fig:cup_clutter_internal}~(left).
As YOLO provides a list of detected object names and their corresponding visual features, we can easily select the object of interest (the cup, in our case) and use its visual features for VS, see Fig.~\ref{fig:cup_clutter_internal} (right). 
Once the cup is selected as the desired object, the robot executes the required motion to position its gripper over the cup. Additionally, in a cluttered scene, YOLO offers us the flexibility of easily changing the target object to any other object detected in the scene.    

\begin{figure}[t!]
\centering
\includegraphics[width=0.48\columnwidth]{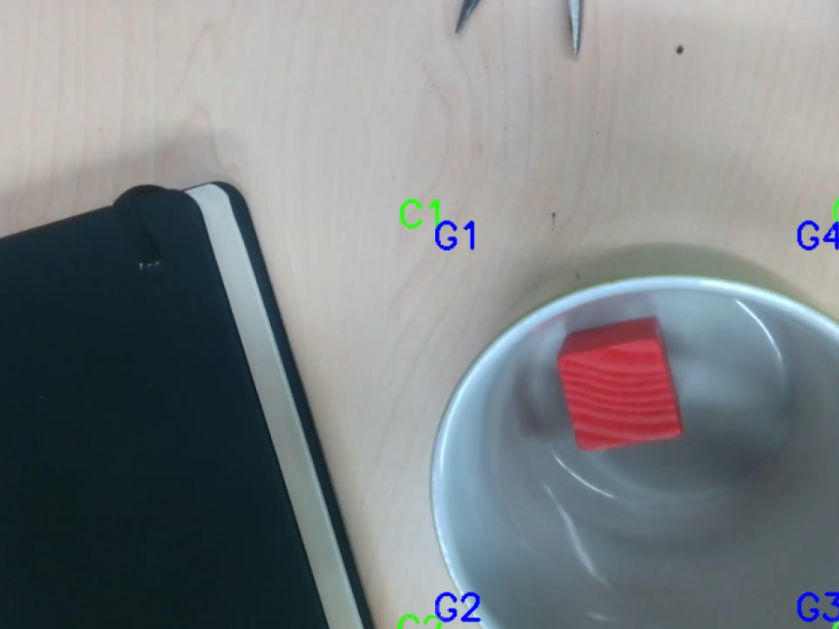}
\hfill
\includegraphics[width=0.48\columnwidth]{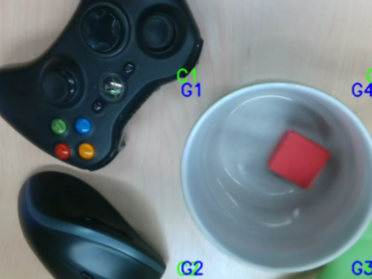}
\caption{Two final image frames captured by the robot camera showing that \revise{\ac{ildvs}} successfully drives the robot to drop the object into the cup in different cluttered environments.
}\label{fig:cup_clutter_internal_examples}    
\end{figure}

With our \revise{\ac{ildvs}} approach, the robot successfully drops the object into the cup in a cluttered setting and also adapts its pose to the different locations of the cup, as shown in Fig.~\ref{fig:cup_cluttered} (note that the locations of the cup and the corresponding final poses of the robot in these snapshots). 
Finally, Fig.~\ref{fig:cup_clutter_internal_examples} shows some views of the object being dropped into the cup placed in different cluttered scenes, as seen from the robot's \revise{end-effector} camera.

\section{Conclusion}\label{sec:conclusion}

In this paper, we have presented \revise{\acf{ildvs}}, a dynamical system-based imitation learning approach for direct visual servoing.
The proposed framework overcomes several limitations of existing approaches. 
\revise{\ac{ildvs}} exploits off-the-shelf deep learning-based perception to extract features from raw camera images, augmented with imitation learning layers that generate complex robot trajectories. 
A key difference from end-to-end learning approaches is that \revise{\ac{ildvs}} exploits a control theoretical framework to ensure convergence to a given target. 
The approach has been extensively evaluated with real robot experiments, and compared with two baselines showing superior performance.    

\section*{Acknowledgements}
Funded by the European Union projects INVERSE (grant agreement No. 101136067) and SERMAS (grant agreement No. 101070351), and by the Swiss State Secretariat for Education, Research and Innovation (SERI) under contract number 22.00247.

Sayantan Auddy is supported by a Doctoral Scholarship from the University of Innsbruck's Support Programme for Young Researchers, awarded by the University of Innsbruck, Vice-Rectorate for Research.

\balance
\bibliographystyle{elsarticle-num}
\bibliography{bibliography.bib}

\end{document}